\documentclass{article}

\PassOptionsToPackage{numbers, sort&compress}{natbib}
\usepackage[final,sglblindworkshop]{neurips_2025}
\workshoptitle{2nd Workshop on Principles of Generative Modeling (PriGM)}
\makeatletter
\renewcommand{\@noticestring}{40th Conference on Neural Information Processing Systems (NeurIPS 2026) Workshop: 2nd Workshop on Principles of Generative Modeling (PriGM).}
\makeatother

\usepackage[utf8]{inputenc}
\usepackage[T1]{fontenc}
\usepackage{hyperref}
\usepackage{url}
\providecommand{\doi}[1]{doi: \href{https://doi.org/#1}{\begingroup\urlstyle{rm}\nolinkurl{#1}\endgroup}}
\usepackage{booktabs}
\usepackage{array,calc}
\newcolumntype{R}[1]{w{r}{\widthof{#1}}}
\newcolumntype{M}[1]{>{\(}w{r}{\widthof{\(#1\)}}<{\)}}
\usepackage{amsfonts}
\usepackage{amsmath}
\usepackage{microtype}
\usepackage{graphicx}
\usepackage{placeins}
\usepackage{orcidlink}
\hypersetup{colorlinks=false,pdfborder={0 0 1},linkbordercolor={1 0 0},citebordercolor={0 1 0},urlbordercolor={0 1 1},pdfauthor={Kevin Lee},pdftitle={Residual Penalties Versus Algebraic Completion In Conditional Diffusion},pdfsubject={},pdfkeywords={}}

\title{Residual Penalties Versus Algebraic Completion\\In Conditional Diffusion}
\author{%
 Kevin Lee{\hypersetup{hidelinks}\orcidlink{0009-0004-0388-9260}} \\
 [4pt]
 University of California, Los Angeles \\
 NASA Jet Propulsion Laboratory \\
 [2pt]
 \texttt{kevinlee69720@g.ucla.edu}
}

\begin{document}
\maketitle

\begin{abstract}
Field components and magnitudes give the same mean squared field strength in each five-minute bin. Averaging the moments across possible reconstructions of a gap can break that equality. Does enforcing it improve conditional diffusion forecasts of a missing hour from observations on either side? In the National Aeronautics and Space Administration's OMNI solar wind data, penalizing disagreement lowers the mean test energy score relative to unpenalized diffusion at the final training step. Choosing the step with the lowest validation score for each fit nearly eliminates the difference. One model computes its loss on five predicted quantities and derives vector variance from them. It has the lowest mean test score for the five quantities jointly, yet its 90\% prediction intervals for the variance of field magnitude cover only 40.5\% of withheld values on average. Clipping to moments attainable by five vectors reduces the mean distance between members enough to worsen both Gaussian baselines' scores. In a two-state hidden Markov model, posterior state probabilities determine the denoising mean. After one step from a hidden path corrupted by strong noise, a second evaluation gives a small mean absolute residual. Decoding each bin separately gives fewer state changes than the model predicts.
\end{abstract}

\section{Introduction}

A strong solar wind magnetic field can have a mean vector near zero when its direction changes. Magnetic energy density depends on mean squared field strength, including fluctuations lost from the mean vector. \citet{wrench2022statistics} estimate second-order structure functions from incomplete magnetic records. Here the question is which combinations of magnetic means and variances are likely during a missing hour, given observations before and after it.\footnote{Code and evaluation data: \url{https://github.com/magnaprog/diffusion-constraints}}

Consider two equally likely constant fields of equal strength pointing in opposite directions for five minutes. Their averaged moments have a zero mean vector, positive mean magnitude, and zero variances. To have zero vector variance, all five vectors would have to be identical. A zero mean would then make all five zero, contradicting the positive mean magnitude. Yet the conditional mean minimizes squared prediction error.

A residual penalty can act on a denoising mean or an approximate sample \citep{bastek2025pidm}. The two opposite constant fields illustrate how penalizing their conditional mean can compete with learning the distribution \citep{baldan2025physics}. An alternative derives some quantities algebraically from learned outputs \citep{beucler2021constraints}. Conditional diffusion samples possible completions of the gap \citep{tashiro2021csdi}. In the solar wind experiment, a penalty encourages agreement between two calculations of mean squared field strength, and a model called G5 derives vector variance from five learned quantities. Clipping already generated samples tests what changes when every bin must be realizable. A two-state hidden Markov model (HMM) gives the probability of each path across the gap. Its denoising mean and path probabilities can be checked separately.

\section{Magnetic moments}

\subsection{Realizability}

Let $B_1,\ldots,B_5$ be the OMNI vector averages for five consecutive minutes from one spacecraft, in geocentric solar ecliptic (GSE) coordinates and nanoteslas (nT). Their vector mean is $\mu$, their mean magnitude is $m$, and the variance of their magnitudes is $v_s$. Vector variance $v_v$ sums the component variances.
\begin{equation}
 \mu=\frac15\sum_j B_j,\quad m=\frac15\sum_j\|B_j\|,\quad
 v_s=\frac15\sum_j(\|B_j\|-m)^2,\quad
 v_v=\frac15\sum_j\|B_j-\mu\|^2.
\end{equation}
The moments $z=(\mu_x,\mu_y,\mu_z,m,v_s,v_v)$ discard the order of the measurements. Mean squared strength is $m^2+v_s$ from the magnitudes and $\|\mu\|^2+v_v$ from the vector components. Their difference gives the residual for the observation rule,
\begin{equation}
 R_{\mathrm{obs}}(z)=m^2+v_s-\|\mu\|^2-v_v=0.
 \label{eq:identity}
\end{equation}
The Physics-Informed Solar-wind Convolutional autoEncoder for Space-weather (PISCES) reconstructs the field vector and its magnitude in separate channels. Its penalty compares their squared strengths minute by minute \citep[Equation~1]{lee2026pisces}. Suppose the field points in opposite directions in successive minutes but stays constant within each. Each minute's mean vector then has the same magnitude as the field. Within a bin, any two nonzero vectors pointing in different directions give $R_{\mathrm{inst}}=m^2-\|\mu\|^2=v_v-v_s>0$. Multiplying all five vectors by the same factor multiplies this residual by the square of that factor. Even an exact reconstruction incurs the instantaneous penalty in this example.

Even a negative variance of field magnitude can satisfy Equation~\ref{eq:identity}. Which moments can five vectors produce? With $r=\|\mu\|$, the necessary and sufficient conditions are
\begin{equation}
 m\ge r,\qquad 0\le v_s\le\frac32m^2+\frac52r^2,\qquad
 v_v=m^2+v_s-r^2.
 \label{eq:support}
\end{equation}
Appendix~\ref{app:support} constructs five vectors for every tuple in this range. Clipping keeps $\mu$, raises $m$ to at least $r$, clips $v_s$ to its permitted range, and calculates $v_v$ from the equality.

These conditions describe an individual reconstruction. A denoiser trained by squared error estimates a conditional mean, where a nonlinear constraint can fail \citep{zhou2025priors}. Taking expectations in Equation~\ref{eq:identity} given the available information $Y$, with finite second moments for $m$ and $\mu$, gives
\begin{equation}
 R_{\mathrm{obs}}(\mathbb{E}[z\mid Y])
 =\operatorname{tr}\operatorname{Cov}(\mu\mid Y)-\operatorname{Var}(m\mid Y).
 \label{eq:conditional}
\end{equation}
Both terms on the right measure uncertainty across possible reconstructions, as in the quadratic correction of \citet[Equation~(26)]{bien2026bias}. The variances $v_s$ and $v_v$ measure fluctuations within a reconstruction. For the two opposite fields, the mean magnitude is the same in both cases. Uncertainty in direction gives a positive residual at the conditional mean. Its sign reverses when uncertainty in mean magnitude exceeds the total variance of the mean field components.

\subsection{Observations}

Hourly OMNI combines field and plasma measurements from several spacecraft \citep{king2005solar}. The modified one-minute product used here \citep{omnimodified2020,nasamodifiedomni} averages the measurements over one-minute intervals of estimated arrival time at the bow shock nose \citep{papitashvili2020omni}. The experiment hides an hour on this arrival-time grid. A single spacecraft measurement can contribute both to the hidden hour and to the surrounding observations.

Each gap has 90 minutes of context on each side. Windows require valid magnetic vectors from one spacecraft every minute. Of 957{,}720 five-minute grid starts whose four hours lie within the training dates and one calendar month, 122{,}309 (12.77\%) qualify. The experiment evaluates forecasts for hours withheld from complete records. Each year from 2005 through 2024 contributes separate training (TRAIN), test (TEST), and validation (VAL) periods with unused dates between them. Normalization uses distinct training bins. Each bin appears in at most one evaluation window (Appendix~\ref{app:quality}).

\section{Comparisons}

\subsection{Conditional diffusion methods}

The unpenalized diffusion model (D) \citep{ho2020ddpm} uses a U-Net to denoise all six quantities in the gap given the measurements on either side. Two methods add a penalty to the same prediction task. The instantaneous penalty (I-S) uses $R_{\mathrm{inst}}$, which can be large even on the measured moments when the field changes direction. The penalty for the observation rule (O-S) subtracts $v_v-v_s$ before squaring, so its residual is zero on measured bins. G5 instead predicts the five quantities sufficient to calculate $v_v$ from Equation~\ref{eq:identity}. This removes both an output and its contribution to the training loss.

For both penalties, S denotes the sample estimate at which the residual is evaluated \citep{bastek2025pidm}. The loss adds the squared residual to the mean squared error (MSE) on the standardized hidden state $u$, with the MSE weight determined by the signal-to-noise ratio (SNR) \citep{hang2023minsnr},
\begin{equation}
 \mathcal L_t=w_t\,\operatorname{MSE}(\hat u_0,u_0)
 +\frac{c_R}{2\Sigma_t}\,\frac1{12}\sum_{k=1}^{12}
 \left(\frac{R(z^*_{0,k})}{B_{\mathrm{ref}}^2}\right)^2.
\end{equation}
For $t>1$, S uses the first prediction $\hat u_0$ in a denoising diffusion implicit model (DDIM) step to index 1 \citep{song2021ddim}, then evaluates the network there. At $t=1$, S uses the first prediction. Both penalties evaluate the resulting estimate $z^*_0$ in physical units. Here $w_t=\min(\operatorname{SNR}_t,5)$ and $\Sigma_t$ is the posterior diffusion variance with $\Sigma_1:=\Sigma_2$. The scale $B_{\mathrm{ref}}^2$ averages $m^2+v_s$ over distinct training bins. Appendix~\ref{app:protocol} gives the cosine schedule \citep{nichol2021improved} and coefficient.

Each fit saves weights at five common steps. Comparing fits at the same step gives equal exposure to training data. Validation can choose different steps for different fits. I-S and O-S each evaluate the denoiser twice on training examples with $t>1$.

\subsection{Evaluation}

The primary energy score compares forecasts of $(\mu_x,\mu_y,\mu_z,m,v_s)$ with the measurements in all twelve hidden bins. On these 60 coordinates standardized on TRAIN, it estimates the mean Euclidean distance from a member to the measured vector minus half the mean distance between distinct members \citep{gneiting2007strictly}. For equal mean distance between members, the ensemble closer to the measurements on average scores better. The estimate is unbiased when members are sampled independently from the forecast distribution. Coverage assesses each quantity and bin separately.

For each fit, raw and clipped forecasts use the earliest steps minimizing their respective validation scores. Deterministic DDIM generates 39 members per window from independent Gaussian noise. For each of the six D and O-S fits, the same step minimizes the raw validation score at 25 and 50 denoiser evaluations per member. TEST uses 50. Configuration and coefficient choices followed inspection of TEST results. Comparisons give equal weight to months within fits and then to fits. Resampling years measures variation across test periods for those fits and ensembles.

Persistence repeats the six values from the last five-minute bin before the gap. Linear interpolation connects the bins on either side. Both give point forecasts. Gaussian interpolation adds a random error vector to the linear forecast, using a Gaussian fitted to interpolation errors on TRAIN. Gaussian regression models mean error given the surrounding observations, derives $v_v$ from the other five quantities, and uses validation to choose its regularization (Appendix~\ref{app:gaussian}).

\begin{table}[!htbp]
\centering
\caption{Forecast accuracy and physical consistency on 480 TEST windows. Months have equal weight within each fit. Diffusion entries average three fits, and each baseline supplies one ensemble per window. For each diffusion fit, raw and clipped predictions use the checkpoints with the lowest respective validation energy scores. Bold marks the lowest energy score in each column, calculated on 60 coordinates standardized on TRAIN. Coverage checks 60 withheld values per window against marginal intervals from 39 ensemble members. For linear interpolation and persistence, the interval is a point and covers an observation when the two values coincide. The last two columns average over 39 raw samples and twelve hidden bins per window. They give mean absolute $R_{\mathrm{obs}}$ (nT$^2$) and the percentage of bins whose mean magnitude or variance of field magnitude falls outside its possible range by more than the numerical tolerance for that inequality.}
\label{tab:main}
\begin{tabular}{lccR{00.0}R{00.0}R{00.000}R{00.0}}
\toprule
 & \multicolumn{2}{c}{Energy score} & \multicolumn{2}{c}{90\% coverage (\%)} & \multicolumn{2}{c}{Raw samples} \\
\cmidrule(lr){2-3}\cmidrule(lr){4-5}\cmidrule(lr){6-7}
Procedure & Raw & Clipped & \multicolumn{1}{c}{Raw} & \multicolumn{1}{c}{Clipped} & \multicolumn{1}{c}{$|R|$} & \multicolumn{1}{c}{\parbox[c]{19mm}{\centering Outside range (\%)}} \\
\midrule
D & 1.888 & 1.869 & 72.8 & 71.0 & 2.956 & 50.4 \\
I-S & 1.882 & 1.876 & 69.4 & 69.6 & 3.732 & 37.8 \\
O-S & 1.885 & 1.866 & 70.0 & 71.1 & 2.680 & 55.7 \\
G5 & \textbf{1.785} & \textbf{1.784} & 75.7 & 75.8 & 0.000 & 48.8 \\
\addlinespace[2pt]
Linear interpolation & 2.393 & 2.393 & 0.0 & 0.0 & 3.459 & 0.0 \\
Persistence & 3.067 & 3.067 & 0.1 & 0.1 & 0.000 & 0.0 \\
Gaussian interpolation & 2.125 & 2.229 & 94.3 & 93.6 & 15.491 & 77.7 \\
Gaussian regression & 1.935 & 2.035 & 94.0 & 93.5 & 0.000 & 74.6 \\
\bottomrule
\end{tabular}
\end{table}

\section{Results}

\begin{figure}[!ht]
 \centering
 \includegraphics[width=\linewidth]{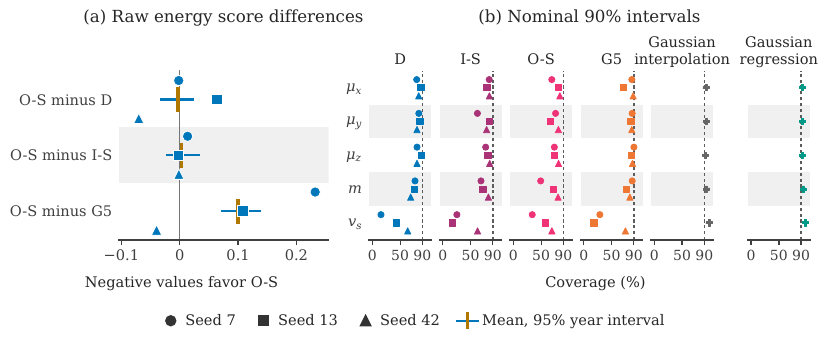}
 \caption{Raw forecasts at the step with the lowest raw validation score for each fit. (a) Each row subtracts the TEST energy score of D, I-S, or G5 from that of O-S at the same seed. Negative differences favor O-S. Gold bars mark the mean of the three seed differences. Horizontal lines give 95\% intervals from resampling years for the same fits and ensembles. (b) Each diffusion symbol shows one fit's coverage for one quantity over twelve hidden positions, with equal month weights. Each Gaussian method supplies one ensemble per window (crosses). Dashed lines mark the nominal 90\% coverage.}
 \label{fig:overview}
\end{figure}

At the final training step the averaging penalty improves the raw TEST energy score over unpenalized diffusion at every seed (Figure~\ref{fig:checkpoints}). Both methods score worse there on average than at the steps chosen by raw validation score, and D loses more accuracy. At each fit's validation choice the mean difference, O-S minus D, is only $-0.0028$ (Figure~\ref{fig:overview}a). Its 95\% interval from resampling years crosses zero. Relative to D, the intervals from resampling individual years or pairs of consecutive years both stay within two percent in either direction. They measure variation across test years for the same three pairs of fits and their ensembles.

At the final step, raising the middle nonzero coefficient tenfold nearly halves the mean absolute residual and shrinks the mean score advantage over D.

\begin{figure}[!htbp]
 \centering
 \includegraphics[width=\linewidth]{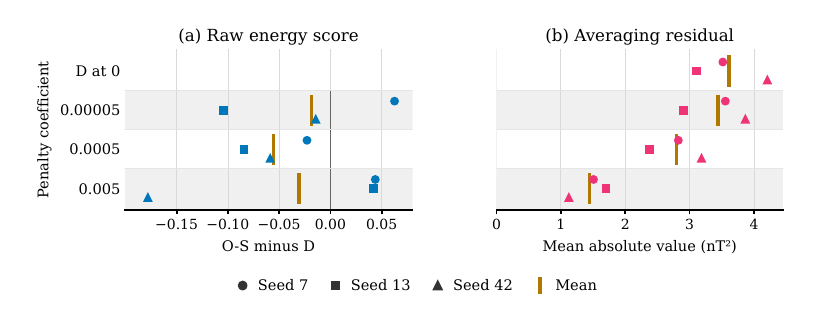}
 \caption{Raw forecasts after 29{,}636 training steps. The rows show unpenalized diffusion (D) and O-S at three penalty coefficients. (a) Each raw TEST energy score minus D's score at the same seed. Negative values favor O-S. The D row is empty because D supplies the reference. (b) Mean absolute averaging residual in nT$^2$, including D at zero penalty. Circles, squares, and triangles mark seeds 7, 13, and 42. Gold vertical bars give their means. Each fit gives TEST months equal weight.}
 \label{fig:fixedfinal}
\end{figure}

At the strongest penalty only one of three fits scores better than D (Figure~\ref{fig:fixedfinal}). At the steps chosen by validation, O-S comes closer to the averaging identity on average than I-S, whose penalty omits the variance difference $v_v-v_s$. I-S has a slightly lower mean raw energy score, although O-S scores lower at two of the three seeds (Table~\ref{tab:main}, Figure~\ref{fig:overview}a).

G5 has the lowest mean energy score when each fit uses the step with the lowest raw validation score, although O-S scores lower at one seed (Figure~\ref{fig:overview}a). Only the five direct predictions enter the score. Deriving vector variance leaves this score unchanged. G5 training changes both the outputs and the loss. Greater mean distance between members accounts for most of G5's score advantage over D.

At seed 13, G5 members are closer together, but their mean distance to the measurements falls enough to lower the score (Figure~\ref{fig:calibration}).

The equality holds for every G5 sample up to rounding. In nearly half its raw bins, the mean magnitude or the variance of field magnitude falls outside the possible range (Table~\ref{tab:main}). Its 90\% prediction intervals for the variance of field magnitude contain only 40.5\% of withheld values on average (Figure~\ref{fig:overview}b). D's coverage is similarly low.

\FloatBarrier

Scoring the same members before and after clipping isolates the effect of the clipping rule (Table~\ref{tab:operators}). In both Gaussian baselines the members move closer to the observations on average. Half the mean distance between members falls by more, so both scores rise. G5 samples move little, and its mean score changes little. For D, one fit changes its preferred training step when validation uses clipped forecasts. Figure~\ref{fig:operator}(b) separates that change from clipping the original ensemble.

For Gaussian regression the loss of coverage is particularly clear in the forecast of hourly magnetic energy density. Each member supplies twelve estimates of $m^2+v_s$ whose mean determines its energy density for the hour. The withheld value uses one vector average per minute (Table~\ref{tab:alternative}). On average, clipping widens the 90\% intervals and raises their lower endpoints. More hourly observations fall below their intervals, and coverage drops from 94.0\% to 67.5\%.

OMNI supplies one observed outcome per gap. In the HMM the conditional probabilities of all paths are known. The first experiment adds noise to a path drawn given the geometry and both endpoint states. At high noise ($t=100$), the second evaluation after one DDIM step gives a small mean absolute residual but a worse energy score than the baseline estimated from independent conditional paths in all 36 cases. When the endpoints agreed, all 2{,}688 decoded paths in this experiment were constant, although the conditional probability is only 18.20\% (Figure~\ref{fig:hmm}). A separate experiment starts from Gaussian noise (Table~\ref{tab:oracle}). Across 288 simulated windows, DDIM with two denoiser evaluations per member scores well above the conditional reference. With 25, the scores are close. Both experiments use the correct denoising mean. In OMNI that mean must also be learned.

\section{Conclusion}

The mean vector, mean magnitude, and variance of field magnitude determine vector variance in a measured bin. Every possible record satisfies this equality, whatever its probability. G5 had the lowest mean primary score, yet its 90\% intervals missed over half the withheld variances of field magnitude. At the steps chosen by validation, O-S lost most of its advantage over D. These comparisons concern gaps imposed in complete OMNI windows, three fits per diffusion configuration, and test data examined during development. Even with the correct denoising mean, decoding after one step from high noise produced too few state changes.

\clearpage
\begin{ack}
This work extends research carried out at the Jet Propulsion Laboratory, California Institute of Technology, under a contract with the National Aeronautics and Space Administration (NASA). Parts of the development and verification were conducted with the Autonomous Research and Engineering System (ARES) agentic platform and its command-line interface, both developed by the author. The author thanks the magnetometer team on the Advanced Composition Explorer (ACE) and the Wind Magnetic Field Investigation team for the magnetic measurements, and Joseph H. King, Natalia E. Papitashvili, and NASA’s Space Physics Data Facility for the OMNI data. The author has no competing interests to declare.
\end{ack}

\bibliography{references}
\bibliographystyle{unsrtnat}
\clearpage
\appendix

\FloatBarrier
\section{Realizable moments}
\label{app:support}

For five vectors with mean vector $\mu$ and mean magnitude $m$, what values can the variance of field magnitude $v_s$ take? Write $a_j=\|B_j\|$ and $r=\|\mu\|$. The triangle inequality gives $r\le m$. For any two vectors, $a_i a_j+B_i\cdot B_j\ge0$, so expanding the squares gives
\[
 2\sum_j a_j^2\le\Bigl(\sum_j a_j\Bigr)^2+\Bigl\|\sum_j B_j\Bigr\|^2
 =25(m^2+r^2).
\]
Since the left side equals $10(m^2+v_s)$, $v_s\le v_{\max}:=3m^2/2+5r^2/2$. For lengths $e=(5(m+r)/2,5(m-r)/2,0,0,0)$, take $B_1=e_1u$ and $B_2=-e_2u$ along a unit vector $u$. Their sum has length $5r$, and the five lengths have variance $v_{\max}$.

To obtain every variance between zero and this maximum, interpolate the lengths from $(m,m,m,m,m)$ to $e$. Set $a_j(\tau)=m+\tau(e_j-m)$ for $0\le\tau\le1$. For $m>0$, append an edge of length $5r$. The six edges can form a closed planar polygon exactly when no edge exceeds half the perimeter, even if some edges have zero length or the polygon lies on a line \citep[Corollary~4.2]{hausmann1997polygon}. The closing edge satisfies this condition because $r\le m$. Each of the other five lengths is at most $5(m+r)/2$ throughout the segment, so the five vectors can have a resultant of length $5r$. Along the segment, $v_s$ is
\[
 \frac15\sum_j\bigl(a_j(\tau)-m\bigr)^2=\tau^2v_{\max}.
\]
For $v_{\max}>0$, set $\tau=\sqrt{v_s/v_{\max}}$ to obtain $v_s$. Choose the directions so that the vector sum has length $5r$, then rotate all five vectors together to align a nonzero sum with $\mu$. When $r=0$, the resultant is zero. When $v_{\max}=0$, all five vectors vanish. In every case, expansion of the vector variance gives $v_v=m^2+v_s-r^2$. This construction realizes all six quantities with the same five vectors.

Equation~\ref{eq:identity} holds for every realizable bin. The conditional mean averages the moments under their distribution given $Y$. Assume finite conditional second moments for $\mu$ and $m$ and finite conditional first moments for $v_s$ and $v_v$. Taking the conditional expectation of Equation~\ref{eq:identity} and substituting
\[
 \begin{aligned}
 \mathbb E[m^2\mid Y]&=\mathbb E[m\mid Y]^2+\operatorname{Var}(m\mid Y),\\
 \mathbb E[\|\mu\|^2\mid Y]&=\|\mathbb E[\mu\mid Y]\|^2+
 \operatorname{tr}\operatorname{Cov}(\mu\mid Y)
 \end{aligned}
\]
gives $R_{\mathrm{obs}}(\mathbb E[z\mid Y])=\operatorname{tr}\operatorname{Cov}(\mu\mid Y)-\operatorname{Var}(m\mid Y)$, as stated in Equation~\ref{eq:conditional}. The conditional variance and covariance terms describe uncertainty across possible bins given $Y$, whereas $v_s$ and $v_v$ describe fluctuations within one bin. The Hessian of $R_{\mathrm{obs}}$ in the coordinate order of $z$ is $\operatorname{diag}(-2,-2,-2,2,0,0)$. Because the residual is quadratic, the Taylor remainder in \citet[Equation~(26)]{bien2026bias} vanishes, and the Hessian gives this correction exactly.

A bin contains five vector averages, one per minute. Their mean squared magnitude is $Q=m^2+v_s=\|\mu\|^2+v_v$. Multiplying $Q$ in nT$^2$ by $10^{-18}/(2\mu_0)$ gives the magnetic energy density in J\,m$^{-3}$ represented by those vectors. Averaging the field over each minute removes finer fluctuations from this energy estimate.

Mean squared strength over a quarter hour can be calculated directly from all fifteen vectors or by averaging $Q$ over the three five-minute bins. Both give each minute equal weight, even when the field changes direction between bins. Across all 97{,}699 eligible intervals beginning on a quarter hour in Coordinated Universal Time (UTC) within the union of training windows, the largest absolute difference was $9.09\times10^{-13}\,\mathrm{nT}^2$. Replacing the vectors in each bin with their mean omits the variance within that bin and underestimates mean squared strength by $0.8943\,\mathrm{nT}^2$ on average.

\FloatBarrier
\section{OMNI window selection}
\label{app:quality}

\FloatBarrier
\subsection{OMNI record}

For each minute, OMNI reports the mean field magnitude and the components of the mean field vector in geocentric solar ecliptic (GSE) coordinates. When the direction varies within a minute, the norm of the mean vector can be smaller than the mean magnitude. The five magnitudes used here are the norms of the five one-minute mean vectors. All six moments then satisfy Equation~\ref{eq:identity} because they come from the same five vectors.

The experiment withholds twelve five-minute bins on OMNI's grid of estimated arrival times at the bow shock nose. To estimate arrival times, OMNI calculates phase front normals from nearby observations or interpolates between normals that pass its checks. For each output minute, OMNI weights each input record by its overlap duration times the fraction of the record within that minute. If a shifted measurement spans the start or end of that hour, it can contribute to minute averages in both a withheld bin and the neighboring observed bin. OMNI also retains both measurements when the later of two observations at one spacecraft has the earlier estimated arrival time. One output minute can then include measurements from different plasma regions \citep[Sections 3a--3c and Appendix 5]{papitashvili2020omni}.

The duration between observing times (DBOT) is negative when the inferred spacecraft observing time decreases from one record to the next on the grid of arrival times. Of 4{,}298{,}363 valid magnetic minutes on training dates with a DBOT value, 555{,}719 (13\%) had negative values. This count includes minutes outside eligible windows. Of the 300{,}392 distinct bins in eligible training windows, 55{,}339 (18\%) contained at least one minute with negative DBOT. For each one-minute magnetic average, OMNI also reports the percentage of contributing measurements with an interpolated phase front normal \citep{nasamodifiedomni}. Variation in the resulting moments can reflect both OMNI processing and changes in the solar wind.

\FloatBarrier
\subsection{Complete windows}

All four hours must fall in one calendar month and come from a single spacecraft, with five consecutive valid magnetic vectors per bin. The calendar fields and magnetic spacecraft identifier must be valid integers. Timestamps must increase strictly, and all three field components must be finite and differ from the OMNI fill value. A window that meets these criteria qualifies regardless of the field amplitude or spikes during the hidden hour.

In each year from 2005 through 2024, TRAIN runs from January 15 through June 30, TEST covers August and September, and VAL runs from November 1 through December 17. The intervening dates are unused. Within each evaluation month, complete windows are packed chronologically without overlap. From $N$ such windows, the sample retains $q=\min(N,q_0)$ at indices $\lfloor(j+0.5)N/q\rfloor$, $j=0,\ldots,q-1$. With $q_0=8$ for VAL and $q_0=12$ for TEST, the index rule spreads the sample across the available windows in each month.

On the five-minute grid, 122{,}309 of the 957{,}720 starts with all four hours on TRAIN dates in the same calendar month met the data criteria. Because the windows overlap, their hidden intervals cover 163{,}988 distinct bins. The number of eligible starts ranged from 2{,}568 in 2014 to 11{,}340 in 2005. Both data availability and solar wind conditions could contribute to this variation.

Even when a spacecraft measures the field, OMNI can have a gap. To propagate a magnetic measurement, OMNI needs the plasma velocity and an estimate of the phase front normal. Extended intervals with little field variation can make that normal difficult to estimate. OMNI has favored Wind since 2020 for data quality and timeliness. Under its earlier rule, OMNI chose one spacecraft for both field and plasma measurements every three hours. The choice depended on magnetic field coverage and the spacecraft's separation from Earth perpendicular to the flow. OMNI kept the previous source in close cases \citep[Appendix 6]{papitashvili2020omni}. Wind supplied every TEST window from 2020 through 2024 except the ACE window on September 29, 2020. Differences in scores by spacecraft can reflect both the observing period and differences in instrumentation and processing.

Although a TRAIN bin may appear in several windows, it contributes only once to normalization. The training mean and population standard deviation for each quantity use all distinct bins in these windows, including the observations on either side of the gap.

\FloatBarrier
\section{Training and evaluation}
\label{app:protocol}

\FloatBarrier
\subsection{Diffusion implementation}

All four diffusion methods train for 31 epochs at each of three seeds. Training records the weights of every fit at the same five steps. Figure~\ref{fig:checkpoints} compares D and O-S at each step and at the step with the lowest validation score for each fit.

The U-Net encoder combines information across progressively longer stretches of the record. Skip connections carry encoder features to the decoder at the same temporal resolution. The companion training code specifies the network dimensions, learning rate, and other AdamW parameters. Evaluation uses the saved weights without exponential averaging.

The diffusion process follows the cosine schedule of \citet[Section~3.2]{nichol2021improved} with 100 noise indices, offset 0.008, and $\beta_t$ capped at 0.999. Here $\beta_t$ is the variance of the noise added at step $t$, and $\bar\alpha_t=\prod_{s=1}^t(1-\beta_s)$. In the forward process, the variance per coordinate of the previous noisy state given the current state and the original sequence is $\beta_t(1-\bar\alpha_{t-1})/(1-\bar\alpha_t)$ \citep[Equations~(6)--(7)]{ho2020ddpm}. The residual penalty takes this variance as $\Sigma_t$ at $t>1$. At $t=1$, the formula gives zero, so the penalty uses $\Sigma_1:=\Sigma_2$. The penalty multiplier is $c_R/(2\Sigma_t)$, with $c_R=0.0005$ for I-S and O-S and zero for D and G5.

The calculations use double precision for the magnetic moments from the measurements and for forecast scores. The network, the residual penalty during training, and the diffusion sampler use single precision. During training, PyTorch samples the noise indices and Gaussian noise on the central processing unit. At evaluation, a 64-bit permuted congruential generator (PCG64) supplies initial noise in six coordinates, of which G5 uses the first five. For the validation scores used to choose checkpoints, each diffusion ensemble contains 39 samples generated with 25 DDIM evaluations per sample. TEST uses 50 evaluations per sample, as does the additional validation comparison in Appendix~\ref{app:checkpoints}. During training and inference, the network receives the measurements outside the gap at every evaluation.

G5's training loss and its evaluation score treat the sixth quantity differently. Omitting $v_v$ reduces the MSE average from 72 hidden coordinates to 60, giving each remaining error more weight. Comparing G5 with D measures the combined effects of this change to the loss and the changes in the network's input and output dimensions. At evaluation, deriving $v_v$ from a given ensemble leaves its primary energy score unchanged because the score already considers only the first five quantities.

\FloatBarrier
\subsection{Residual estimates}

The residual penalty takes an estimate of the clean values in the gap. M takes the first denoising prediction. S uses this prediction for a deterministic DDIM step from $t$ to index 1 \citep[Equation~(12)]{song2021ddim}, then evaluates the denoiser there. At $t=1$, S reuses the first prediction and equals M. Both procedures convert the estimate to physical units before calculating the residual.

For a given context, S is a deterministic function of the noisy input. At the second input, the ideal denoiser returns the conditional mean under forward noising at index 1. Reaching the second input by a DDIM step can change what it implies about the original sequence, including its conditional mean. Appendix~\ref{app:oracle} measures the consequences for a known path distribution.

\FloatBarrier
\subsection{Interpolation baselines}

Linear interpolation and persistence each predict a single sequence for the twelve hidden bins. Linear interpolation assigns weight $(i-17)/13$ to bin 30 at hidden index $i=18,\ldots,29$, and the complementary weight to bin 17, for every coordinate. Persistence repeats the six values from the bin immediately before the gap (bin 17) throughout the missing hour. For either method, the ensemble members agree at each bin and coordinate, so the prediction interval has zero width.

Gaussian interpolation models the errors in the linear interpolant across positions and quantities. Each TRAIN window contributes one vector of 72 standardized errors to a common mean and covariance, with equal weight even where windows overlap. For $N$ training windows the empirical covariance $S$ has divisor $N$. With shrinkage $\rho=0$, the forecast covariance is $S+10^{-6}I$. One ensemble member is the sum of the standardized interpolant and a random error vector from this Gaussian distribution. The interpolant depends on the six summaries in each bin bordering the gap. The error distribution is the same for every window.

\FloatBarrier
\subsection{Gaussian regression}
\label{app:gaussian}

Gaussian regression also models interpolation errors, but lets their mean depend on the entire observed context. The design matrix $X$ contains 216 standardized available coordinates per window. The response $Y$ contains the 60 differences between the primary measurements inside the gap and their linear interpolants, each divided by the TRAIN standard deviation of the corresponding quantity. The coefficients $W$ and unpenalized intercept $b$ minimize $\|Y-XW-b\|_F^2/N+\lambda\|W\|_F^2$, giving every eligible training window equal weight.

For each ridge coefficient $\lambda\in\{0.001,0.1,1\}$, the fit supplies the mean prediction and an empirical covariance $S$ of the remaining errors, with divisor $N$. Shrinkage $\rho\in\{0,0.5,1\}$ gives covariance $(1-\rho)S+\rho\operatorname{diag}(S)+10^{-6}I$. For each combination, the 60 primary coordinates have a Gaussian distribution whose mean varies with the observations outside the gap. Its positive definite covariance is the same for every window. Computing $v_v$ from the moment equation extends each sample to all six quantities. Because $v_v$ depends quadratically on the Gaussian coordinates, the six quantities together have a non-Gaussian distribution. Every sample satisfies the equality, although $m$ or $v_s$ can fall outside the range in Equation~\ref{eq:support}.

For each VAL window, all nine parameter combinations start from the same 39 vectors of standard Gaussian noise. Each validation month has equal weight in the mean energy score. The raw and clipped scores are minimized separately. If scores are equal, the smaller $\lambda$ takes precedence, followed by the smaller $\rho$. Both selections give $\lambda=0.1$ and $\rho=0$, retaining the full residual covariance. These parameter choices give one ensemble per TEST window.

Gaussian regression and Gaussian interpolation differ in their inputs, their models for the conditional mean, their residual covariance estimates, and the ensembles sampled for evaluation. Their scores compare the two procedures as a whole. Appendix~\ref{app:extension} examines the forecast and the effect of clipping.

\FloatBarrier
\subsection{Assessment}

The primary energy score compares forecasts across all five primary quantities and twelve hidden bins together. Scoring the three vector components alone isolates the vector forecast, while scoring all six quantities also assesses the derived or generated vector variance. The continuous ranked probability score (CRPS), interval coverage, and interval width assess each quantity separately. A wide interval can contain the observed value even when the forecast is imprecise.

For a given fit, time, and quantity, the 90\% interval runs from the second to the thirty-eighth of 39 ordered samples. The 50\% interval uses ranks 10 and 30. When the members and observation are exchangeable and all 40 values are distinct, the observation is equally likely to occupy each rank. The probability of falling between members at ranks $l$ and $h$ is then $(h-l)/40$. Independent draws from the same continuous conditional law satisfy these conditions. In the discrete HMM an observation can equal an endpoint. Including these ties can raise coverage above the nominal level.

Table~\ref{tab:geometry} reports vector increment errors at both edges of the gap and compares predicted and observed mean increment magnitudes inside it. Even linear interpolation changes the mean vector by $(\mu_{30}-\mu_{17})/13$ across each end of the gap. A sampled interpolation error can change the first or last predicted vector. Within each bin, $R_{\mathrm{obs}}$ measures the discrepancy between the two estimates of mean squared field strength. The inequality failure rate is the fraction of generated bins whose $m$ or $v_s$ falls outside the ranges in Equation~\ref{eq:support} by more than the rounding tolerance.

\FloatBarrier
\subsection{Criteria for comparing forecasts}

The TEST score at each fit's validation minimum measures the performance of a procedure that includes checkpoint selection. Its interpretation is retrospective because the choice of configuration and penalty coefficient followed inspection of TEST results.

Only D at seed 7 has its lowest raw and clipped validation scores at different training steps (Table~\ref{tab:seeds}). Choosing the checkpoint by its clipped validation score lowers D's mean clipped TEST score relative to clipping at the raw validation minimum. Figure~\ref{fig:operator}(b) separates the effect of changing the training step from that of clipping the predictions at the original step.

Raw O-S meets the superiority criterion only if it outperforms raw D, raw G5, clipped D, and clipped G5. Each clipped method uses its own checkpoint with the lowest clipped validation score for each seed. For each comparison, the O-S mean energy score must be at least two percent lower, every seed difference must favor O-S, and both temporal intervals must have upper endpoints below zero. The first four rows of Table~\ref{tab:contrasts} show where the criterion fails. Even against raw D, one seed favors D, and both intervals cross zero.

The separate tolerance criterion compares raw O-S with raw D. It holds when all endpoints of both relative intervals fall strictly between $-0.02$ and $0.02$, with a positive D denominator in every resample. Different predictive distributions can meet this tolerance on their mean energy scores.

The intervals in Table~\ref{tab:contrasts} come from resampling individual years or groups of two consecutive years (Appendix~\ref{app:details}). Each interval measures how a single comparison varies across test periods for the same fits and ensembles. Uncertainty across training runs and simultaneous coverage across comparisons are outside its scope.

\FloatBarrier
\subsection{Training stage}
\label{app:checkpoints}

\begin{figure}[!ht]
 \centering
 \includegraphics[width=\linewidth]{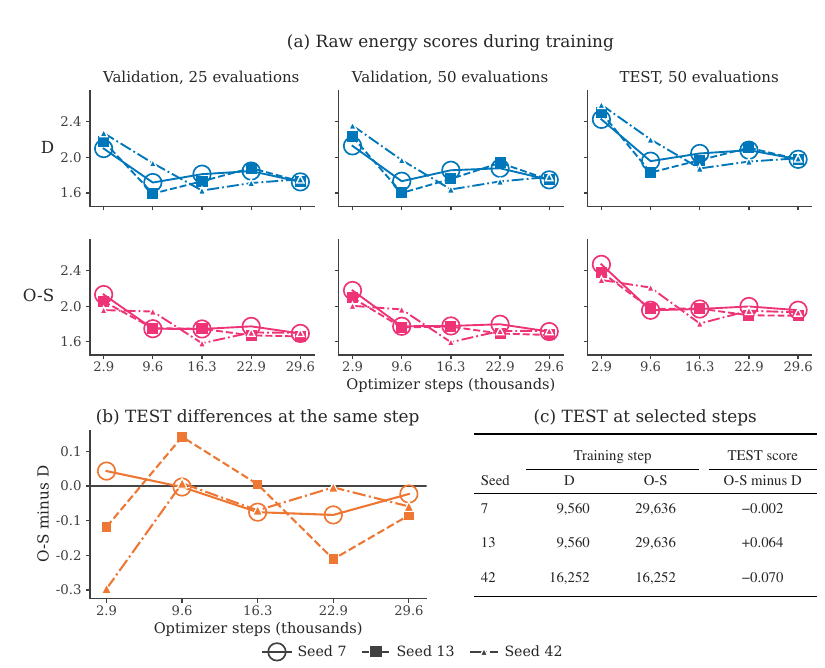}
 \caption{(a) Raw energy scores at five optimizer steps for D (top) and O-S (bottom). The validation columns use 25 and 50 denoiser evaluations per sample. TEST uses 50 evaluations per sample on different windows. Circles, squares, and triangles mark training seeds 7, 13, and 42, respectively. (b) Difference in raw TEST energy score (O-S minus D) at the same optimizer step and training seed. Values below zero favor O-S. (c) The table gives the step with the lowest validation score for each fit and the TEST score difference when each method uses its own selection. These steps differ between D and O-S at seeds 7 and 13. Both validation columns give the same choices. Scores give months equal weight within each fit and use 39 members per ensemble. Checkpoint selection preceded these additional TEST comparisons. The model configuration was chosen after inspection of TEST results.}
 \label{fig:checkpoints}
\end{figure}

Figure~\ref{fig:checkpoints}a follows the raw energy score of each D and O-S fit across five training steps. At each checkpoint, the two validation runs use the same network weights, windows, and initial noise, with either 25 or 50 denoiser evaluations per sample. After checkpoint selection on validation, the TEST comparison at 50 evaluations per sample reused existing forecasts and added forecasts at the remaining checkpoints. The raw validation score was higher with 50 evaluations than with 25 at every checkpoint in Figure~\ref{fig:checkpoints}a. For each of the six D and O-S fits, raw validation scores are lowest at the same step with 25 or 50 evaluations.

On TEST, O-S scores lower than D on average across seeds at four of the five common training steps (Figure~\ref{fig:checkpoints}b). At each of the first three steps, D scores lower for some seeds and O-S for others. O-S scores lower for all three seeds at both of the last two steps. D has its lowest validation score before the final step at every seed. For O-S, the minimum occurs at the final step for seeds 7 and 13 (Figure~\ref{fig:checkpoints}c). From each fit's validation minimum to the final step, the mean TEST score rises by 0.0945 for D and 0.0419 for O-S. The mean TEST score difference, O-S minus D, shifts from $-0.0028$ at the validation minima to $-0.0554$ at the final step.

At seed 7, D's raw validation score at the final checkpoint exceeds its minimum at step 9{,}560 by 0.0073 with 25 evaluations per sample and 0.0156 with 50. Resampling the twenty validation years with both months of each year kept together shows how often a different training step has the lowest score. Step 9{,}560 has the lowest score in 61.0\% of 10{,}000 resamples with 25 evaluations and 71.5\% with 50. At each evaluation count, the other resamples favor the final step. For D at seeds 13 and 42, every resample gives the same step as the full validation set at both evaluation counts.

In each resample of calendar years, one TEST comparison uses the checkpoints chosen on the full validation set. The other repeats checkpoint selection on the validation months from those same years. Each year contributes its validation and test months the same number of times across all fits. Choosing the step with the lowest validation score at 50 evaluations per sample, with ties going to the earlier step, changes at least one fit's checkpoint in 4{,}264 of 10{,}000 resamples. The 95\% percentile interval for the mean TEST score difference, O-S minus D, is $[-0.0346,0.0242]$ at the original checkpoints and $[-0.0336,0.0246]$ at the steps chosen in each resample. Within each resample, subtracting the TEST contrast at the original checkpoints from the contrast after repeating the choice gives the change for those same TEST years. The 2.5th and 97.5th percentiles of these changes are $-0.0206$ and $0.0145$. The two intervals for O-S minus D have similar endpoints and both include zero, although choosing checkpoints again changes the contrast in some resamples. For a given fit, window, and checkpoint, both calculations reuse the same forecasts. Dependence between years can affect the coverage of these intervals.

\FloatBarrier
\section{Hidden Markov model}
\label{app:oracle}

\subsection{Conditional paths}

The magnetic identity holds within a bin. To test whether it can also distinguish plausible paths across a gap, consider a process with two possible states at each bin. The model gives the probability of each hidden path given the geometry and endpoints, and the denoising mean for each noisy input. Each state, $z_0$ or $z_1$, contains the magnetic moments of five vectors and satisfies Equation~\ref{eq:identity}. For $\alpha\in\{0,1,2\}$, let $c_j=j-3$ for $j=1,\ldots,5$ and $d_\alpha=\sqrt{1+2\alpha^2}$. The vectors are
\begin{equation}
 B_j^{(0)}=\frac{(1,\alpha c_j,0)}{d_\alpha},\qquad
 B_j^{(1)}=\frac{2(1+2\alpha c_j,\,2-2\alpha c_j,\,2+\alpha c_j)}{3d_\alpha}.
 \label{eq:toystates}
\end{equation}
Their mean squared magnitudes are $Q_0=1$ and $Q_1=4$ in HMM units. Each state has the same $Q$ at all three geometries, although the variance of the five magnitudes changes with $\alpha$. Standardization weights the six state and geometry combinations equally, giving $B_{\mathrm{ref}}^2=2.5$.

For a path with parameter $p\in\{0.1,0.25,0.5\}$, the state changes with probability $p$ at each step. The transition matrix is $T_p=\bigl(\begin{smallmatrix}1-p&p\\p&1-p\end{smallmatrix}\bigr)$. There are thirteen transitions from the observed left state $a$, through twelve hidden states, to the observed right state $b$. Under the stationary distribution, the initial state $a$ takes either value with probability $1/2$. Given $a$ and $p$, the right endpoint has probability $\Pr(b\mid a,p)=(T_p^{13})_{ab}$, so observing both endpoints gives information about $p$. For fixed $p$, this chain conditioned to start at $a$ and end at $b$ after thirteen transitions is a Markov bridge.

Each simulated window of 48 bins repeats $z_a$ in the 18 bins to the left of the gap and $z_b$ in the 18 to the right. The observed bins determine the geometry and both endpoint states. Together these form the context $\mathcal C=(\alpha,a,b)$ available to every method. Only the thirteen transitions between $a$ and $b$ enter the likelihood for $p$.

The three switching probabilities have equal prior weights $\pi_p=1/3$. For a hidden sequence $\xi=(\xi_1,\ldots,\xi_{12})$, define $\xi_0=a$, $\xi_{13}=b$, $w_p(\xi)=\prod_{j=0}^{12}(T_p)_{\xi_j\xi_{j+1}}$, and $Z_p=(T_p^{13})_{ab}$. The endpoint likelihood $Z_p$ gives
\[
 \Pr(p\mid\mathcal C)=\frac{\pi_p Z_p}{\sum_{p'}\pi_{p'}Z_{p'}},\qquad
 \Pr(\xi\mid\mathcal C)=
 \frac{\sum_p\pi_p w_p(\xi)}{\sum_p\pi_p Z_p}.
\]
To sample a path, first choose $p$ with probability $\Pr(p\mid\mathcal C)$, then sample $\xi$ with probability $w_p(\xi)/Z_p$. One value of $p$ governs all thirteen transitions. Sampling the hidden states together preserves their dependence given the endpoints.

\subsection{Denoising by posterior averaging}

A noisy input changes the relative probabilities of the paths compatible with the endpoints. Let $U_0$ be the standardized hidden path. The forward process gives $U_t=a_tU_0+s_t\epsilon$, with $a_t=\sqrt{\bar\alpha_t}$, $s_t=\sqrt{1-\bar\alpha_t}$, and independent standard Gaussian $\epsilon$. The denoiser that minimizes squared error is the conditional mean $D_t(u,\mathcal C)=\mathbb E[U_0\mid U_t=u,\mathcal C]$. Writing $\widetilde z_k$ for the six standardized coordinates of state $k$, Bayes' rule gives
\[
 \Pr(\xi,p\mid U_t=u,\mathcal C)\ \propto
 \pi_p w_p(\xi)\prod_{j=1}^{12}
 \varphi_6\!\left(u_j;a_t\widetilde z_{\xi_j},s_t^2 I_6\right).
\]
Here $\varphi_6(\cdot;\eta,V)$ is the Gaussian density in six dimensions with mean $\eta$ and covariance $V$. For each $p$, forward and backward recursions calculate the state probabilities and the likelihood of the noisy input. The denoiser averages these probabilities over $p$ with posterior weights $\Pr(p\mid U_t=u,\mathcal C)$, then uses them to weight the two standardized state vectors in each bin.

Let $q$ be the posterior probability of state 1 in a given bin, and define $\kappa_\alpha=[1+2\sqrt{1+\alpha^2}+2\sqrt{1+4\alpha^2}]/5$. In physical coordinates the residual at the posterior mean is
\begin{equation}
 R_{\mathrm{obs}}\bigl((1-q)z_0+qz_1\bigr)
 =q(1-q)\frac{11/3-\kappa_\alpha^2}{1+2\alpha^2}.
 \label{eq:toymean}
\end{equation}
Averaging the residual over the two possible states gives zero for every $q$. Evaluating it at their mean instead gives a positive residual for $\alpha=0,1$ and a negative one for $\alpha=2$ whenever $0<q<1$. In all three geometries a squared residual penalty assigns a cost to the estimate that minimizes squared denoising error. This is the distinction between residuals at a mean and at individual outcomes discussed by \citet{bastek2025pidm}.

\subsection{Two denoiser evaluations}

For $t>1$, S takes the DDIM step $F_t$ to noise level 1 without drawing fresh noise, then applies the denoiser again to obtain $S_t$. Omitting $\mathcal C$ from the notation gives
\[
 F_t(u)=a_1D_t(u)+\frac{s_1}{s_t}\bigl[u-a_tD_t(u)\bigr],\qquad
 S_t(u)=D_1\bigl(F_t(u)\bigr).
\]
At $t=1$, S uses $D_1(u)$ directly. The residual calculation first converts the estimate back to physical units. For $t>1$, S passes $F_t(U_t)$ to the denoiser at noise level 1. At that level the forward process gives $U_1=a_1U_0+s_1\epsilon$. Even for the same clean path $U_0$, the two inputs $F_t(U_t)$ and $U_1$ can have different distributions.

Does the DDIM step discard information about the clean path? By the law of iterated expectations,
\[
 \mathbb E[U_0\mid F_t(U_t),\mathcal C]
 =\mathbb E[D_t(U_t,\mathcal C)\mid F_t(U_t),\mathcal C].
\]
For $t>1$ and a fixed context, distinct noisy inputs give distinct values of $F_t$ before rounding. For a given context, the derivative formula for a conditional mean under Gaussian noise from \citet{dytso2023conditional} gives the Jacobian of $D_t$ as $(a_t/s_t^2)\operatorname{Cov}(U_0\mid U_t=u,\mathcal C)$. This matrix is positive semidefinite, so $D_t$ is monotone. For $t>1$, write $F_t(u)=cu+dD_t(u)$, where $c=s_1/s_t>0$ and $d=a_1-a_ts_1/s_t>0$. Then $(F_t(u)-F_t(v))^\top(u-v)\ge c\|u-v\|^2$. The positive lower bound for distinct inputs makes the step invertible on its image. Conditioning on $F_t(U_t)$ is equivalent to conditioning on $U_t$, with conditional mean $D_t(U_t,\mathcal C)$. The second call instead evaluates $D_1(F_t(U_t),\mathcal C)$ under the forward noise model at level 1, which can give a different mean. On this image the inverse can amplify distances by at most $1/c$. Rounding can still merge nearby outputs.

\subsection{Denoising at four noise levels}

\begin{figure}[ht]
 \centering
 \includegraphics[width=\linewidth]{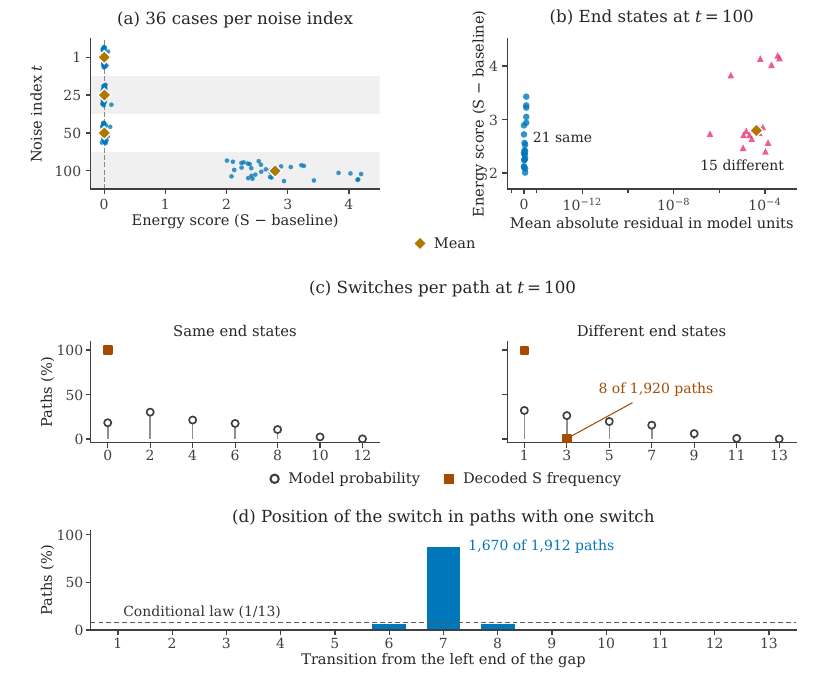}
\caption{Scores, residuals, and state changes when the hidden Markov model supplies the denoising mean. The 36 cases span 12 observable contexts. Each case has 128 S estimates and 128 independent paths from the conditional law. (a) S energy score against those paths minus the baseline score. The baseline is half the mean distance over all pairs of the 128 independently sampled paths. (b) Each point shows one case's mean absolute residual and its score difference at $t=100$. The end states are the same for blue circles and different for pink triangles. The residual axis is linear through $10^{-14}$ and logarithmic above it. Gold diamonds show means over all 36 cases at each noise index in (a) and on both axes in (b). (c) Switch counts at $t=100$ include both transitions at the ends of the gap. The count is even for the same end states and odd for different end states. Filled squares give decoded S percentages. An absent square means zero decoded paths. Open circles show the probability of each count given the end states. (d) Among decoded paths with one switch at $t=100$, bars show the percentage switching at each transition. The dashed line gives the probability under the conditional law given one switch.}
 \label{fig:hmm}
\end{figure}

Figure~\ref{fig:hmm} tests whether S reproduces the conditional distribution of hidden state sequences across repeated noisy inputs. Within each case, the geometry and both observed endpoint states stay the same. At low noise, the posterior probability can concentrate on a single sequence.

The sampling comparison includes 36 cases, the first four evaluation windows in each of the nine combinations of geometry and generating switching probability. The generating probability groups the windows for evaluation but is hidden from both samplers. Both methods receive the same twelve combinations of geometry and endpoint states across the 36 cases. For each case, the experiment samples 128 new paths from the conditional mixture over $p$ and adds independent forward noise to each. S estimates each clean path from its noisy version, the context, and the noise level. A separate random stream supplies 128 paths from the same conditional law to estimate the baseline score. The averages at every noise level give each case equal weight, so repeated contexts count once per case.

At high noise, the residual and the path score give different answers. At $t=100$, the S score exceeds the baseline estimated from independent paths in all 36 cases (means 6.958 and 4.164), although mean absolute $R_{\mathrm{obs}}$ is only $4.17\times10^{-5}$ in HMM units. Mean absolute $R_{\mathrm{obs}}$ is near roundoff at $t=1$ and $25$ (about $6\times10^{-16}$ in HMM units) and reaches $6.94\times10^{-11}$ at $t=50$. When the end states agree, the residuals are near zero and the score differences are large (Figure~\ref{fig:hmm}b). Equation~\ref{eq:toymean} shows how this can happen. A bin estimate near either state has a small residual, regardless of how often the path switches.

Figure~\ref{fig:hmm}(c) decodes a bin to state 1 if the second denoiser call assigns that state a posterior probability above $1/2$, and to state 0 otherwise. Switches are counted from the observed left state through the twelve hidden bins to the observed right state. Only this count and the switch locations use decoded states. Scores and residuals use the continuous estimates. In the 21 cases with the same end states, all 2{,}688 decoded paths stay in that state throughout the gap. Given those endpoints, the probability of a constant path is 18.20\%. In the 15 cases with different end states, 99.58\% of the decoded paths have just one switch, compared with a conditional probability of 32.03\%. The other eight decoded paths have three switches.

Even conditional on one switch, its location is distorted. Among the decoded paths with one switch, 87.34\% switch at the seventh transition. Given the initial state and switching probability $p$, each path with one switch has probability $p(1-p)^{12}$ regardless of where the switch occurs (Figure~\ref{fig:hmm}d). Under the HMM law, conditional on one switch, each of the thirteen positions has probability $1/13$. For different end states, choosing the more probable state at each bin from the endpoints alone gives a single switch at the seventh transition. The concentration at the middle is consistent with that marginal decision.

\FloatBarrier
\FloatBarrier

\subsection{Full DDIM sampling}

Figure~\ref{fig:hmm} starts with a hidden path and adds noise to it. Full DDIM sampling starts from Gaussian noise and generates a path for the observed endpoints and geometry. Table~\ref{tab:oracle} scores these forecasts against the hidden values in all 288 simulated windows, including those that supplied the figure contexts. For each $p$, the three geometries share the same 32 hidden state sequences.

Summing over all possible paths gives the score of the conditional law for each window. Direct sampling estimates that score with a finite ensemble. The DDIM runs start from the same Gaussian samples and differ in the noise levels at which they evaluate the denoiser. With two denoiser evaluations per sample, the score is well above both benchmarks. At 25 evaluations the score is close to both and changes little at 50 or 100 (Table~\ref{tab:oracle}).

A close energy score still allows a difference in the possible outputs. At the last denoising step, both states have positive posterior probability for any finite noisy input. Before rounding, the returned mean falls strictly between the two state vectors at each bin. The HMM assigns probability only to paths that take one of those state values at every bin.

\begin{table}[h]
\centering
\caption{Energy score on 288 simulated windows from the hidden Markov model. The left endpoint takes either state with probability $1/2$ under the stationary distribution. The observed bins on each side repeat their endpoint state. The model score sums over all 4{,}096 hidden paths and averages over the unknown switching probability. Direct sampling follows the same conditional law. Each DDIM call evaluates the denoiser once. Direct sampling and DDIM each generate 39 paths per window. All rows compare forecasts with the same hidden values. The ensemble scores subtract half the mean distance between distinct members (Appendix~\ref{app:scoreformula}).}
\label{tab:oracle}
\begin{tabular}{@{}lcR{000}c@{}}
\toprule
Procedure & Samples per window & \multicolumn{1}{c}{Calls per sample} & Energy score \\
\midrule
Conditional law, all paths & --- & \multicolumn{1}{c}{---} & 4.2207 \\
Direct sampling & 39 & 0 & 4.2287 \\
DDIM & 39 & 2 & 7.2183 \\
DDIM & 39 & 25 & 4.2591 \\
DDIM & 39 & 50 & 4.2586 \\
DDIM & 39 & 100 & 4.2567 \\
\bottomrule
\end{tabular}
\end{table}

\FloatBarrier
\subsection{Diffusion on the HMM}

The preceding comparisons give the sampler the HMM denoising mean. Fitting a denoiser to the same process tests whether the choice of training penalty affects its forecasts. All fits train on the same 4{,}096 windows spanning every combination of geometry and switching probability, with the same U-Net architecture, optimizer, and diffusion schedule (Appendix~\ref{app:protocol}). Table~\ref{tab:learnedtoy} compares nine fits, each sampled with 50 DDIM evaluations.

The clearest difference is which residual enters the loss. At this training seed, O-M and O-S include the variances in the penalty and score close to D at both coefficients. Omitting the variances raises the score for both M and S. I-S at $c_R=0.005$ has a wider mean 90\% interval than D and lower coverage. At each of $\alpha=1$ and $\alpha=2$, this fit's intervals for $\mu_x$ and $m$ each cover all 539 bins in state 0 and none of the 613 bins in state 1. The two geometries reuse the same hidden paths.

\begin{table}[ht]
\centering
\caption{Forecasts from diffusion models trained on the hidden Markov model (HMM). The left endpoint takes either state with probability $1/2$ under the stationary distribution. The observed bins on each side repeat their endpoint state. Every fit trains for 5{,}000 steps at seed 7 and generates 39 paths for each of the same 288 simulated windows. Bold marks the lowest observed energy score among the nine fits. Raw $|R|$ is mean absolute $R_{\mathrm{obs}}$ in HMM units. A generated bin counts once if its mean magnitude or magnitude variance falls outside the range attainable with five vectors by more than the rounding tolerance. The equality residual is assessed separately. Coverage and width average the five primary coordinates and twelve hidden positions. The endpoints of each 90\% prediction interval are the second and 38th of the 39 ordered raw samples. Width is the difference between those endpoints in standardized coordinates. For direct conditional sampling, the 90\% intervals give 100\% coverage on these windows, counting ties at either endpoint.}
\label{tab:learnedtoy}
\begin{tabular}{llccccc}
\toprule
 & & \multicolumn{5}{c}{Raw predictions} \\
\cmidrule(lr){3-7}
Procedure & \multicolumn{1}{c}{$c_R$} & \parbox[c]{14mm}{\centering Energy score} & $|R|$ & \parbox[c]{19mm}{\centering Outside range (\%)} & \parbox[c]{18mm}{\centering 90\% coverage (\%)} & \parbox[c]{13mm}{\centering 90\% width} \\
\midrule
D & 0 & 4.2607 & 0.0126 & 27.5 & 94.5 & 1.4833 \\
I-M & 0.0005 & 4.3283 & 0.2861 & 26.6 & 84.3 & 1.4939 \\
I-M & 0.005 & 4.6021 & 0.7050 & 26.6 & 81.2 & 1.5625 \\
I-S & 0.0005 & 4.4991 & 0.6262 & 26.0 & 77.5 & 1.5396 \\
I-S & 0.005 & 4.8020 & 1.0086 & 29.2 & 74.6 & 1.6499 \\
O-M & 0.0005 & 4.2604 & 0.0122 & 27.8 & 94.7 & 1.4835 \\
O-M & 0.005 & \textbf{4.2544} & 0.0095 & 27.4 & 95.3 & 1.4854 \\
O-S & 0.0005 & 4.2593 & 0.0120 & 27.8 & 94.8 & 1.4836 \\
O-S & 0.005 & 4.2551 & 0.0094 & 27.4 & 95.9 & 1.4855 \\
\bottomrule
\end{tabular}
\end{table}

Discreteness matters even for a forecast from the true law. At a given geometry each coordinate takes at most two values. An interval that reaches both covers every possible value of that coordinate. A value equal to either endpoint counts as covered. On these windows, the 50\% intervals from direct sampling cover 97.49\% of the hidden coordinate values. The mean widths for direct sampling are 1.4581 in standardized coordinates for the 90\% intervals and 1.3485 for the 50\% intervals.

Figure~\ref{fig:hmm} scores the estimates from S itself. In O-S, S supplies the estimate for the penalty during training. The forecast score in Table~\ref{tab:learnedtoy} comes from paths the fitted model generates with the same DDIM schedule as D.

\FloatBarrier
\section{Scores and coverage}
\label{app:details}

\FloatBarrier
\subsection{Forecasts and scores}
\label{app:scoreformula}

Each ensemble member predicts five magnetic quantities in each of the twelve missing bins. An error in a vector component has units of nT, whereas an error in the variance of field magnitude has units of nT$^2$. Centering and scaling each quantity on TRAIN puts all 60 coordinates on a common scale. Let $X$ and $X'$ be independent and identically distributed (iid) draws from the forecast law $F$ conditional on the available measurements. Let $y$ contain the 60 observations withheld for evaluation. The population energy score (ES) compares two expected Euclidean distances
\[
 \operatorname{ES}(F,y)=\mathbb{E}\|X-y\|-\frac12\mathbb{E}\|X-X'\|,
 \qquad X,X'\stackrel{\mathrm{iid}}{\sim}F.
\]
If the forecast and outcome laws have finite first norm moments, the true conditional law minimizes the expected energy score \citep{gneiting2007strictly}. Each TEST window supplies one withheld vector, and the reported score averages across observed contexts. A small observed difference between models can be consistent with either ordering of their expected scores.

For $K\geq 2$ independent ensemble members sampled from the same forecast law, define
\[
 a_K=\frac1K\sum_{k=1}^K\|X_k-y\|,\qquad
 b_K=\frac{1}{K(K-1)}\sum_{k\ne l}\|X_k-X_l\|.
\]
For iid members, $\widehat{\operatorname{ES}}_U=a_K-b_K/2$ estimates the population score without bias. Using the empirical distribution of the members changes the weight on distances between members. Drawing two member indices independently and uniformly gives probability $1/K^2$ to each $(k,l)$. The $K$ cases with $k=l$ contribute zero distance. The empirical distribution's score is $\widehat{\operatorname{ES}}_U+b_K/(2K)$. For a scalar with finite first absolute moment, absolute distances give the fair ensemble CRPS \citep{ferro2014fair}.

Pooling all 117 diffusion members would include distances between predictions from different fits. Each fit contributes a score and prediction intervals from its own 39 members (Table~\ref{tab:seeds}). Raw and clipped forecasts each use the checkpoint with the lowest validation energy score for that form. Scores and coverage average windows within months, months equally, and then fits.

Gaussian interpolation and Gaussian regression each contribute one ensemble per window. Gaussian interpolation uses the same error distribution for all windows. In Gaussian regression, the mean interpolation error depends on the observations around the gap (Appendix~\ref{app:gaussian}).

\FloatBarrier
\subsection{Differences across fits and years}

How much does the choice of TEST years matter for the same fitted models? In each of 10{,}000 resamples, the two methods in a comparison receive the same TEST years (Table~\ref{tab:contrasts}). One scheme samples twenty years with replacement. A second keeps consecutive years together, sampling ten times with replacement from ten nonoverlapping groups, 2005--2006 through 2023--2024. Each occurrence of a year contributes its months again.

In each resample, the relative difference is the mean O-S minus D score difference divided by the mean D score for the same years. The relative 95\% intervals run from $-1.68\%$ to $1.40\%$ for individual years and from $-1.68\%$ to $1.30\%$ for groups of two years. Both intervals meet the criterion for O-S relative to D because all four endpoints fall strictly between $-2\%$ and $2\%$. In score units, both resampling schemes give intervals that cross zero for O-S minus I-S and I-S minus D (Table~\ref{tab:contrasts}). With only twenty years or ten groups, the intervals can miss their intended coverage. Applying them to other solar cycles assumes temporal stability, and retraining would introduce variation beyond the years resampled here.

\begin{table}[ht]
\centering
\caption{Training steps and primary forecast results by run. The step columns give the checkpoints with the lowest raw and clipped validation energy scores, respectively. The two steps differ only for D at seed 7. Energy scores use standardized coordinates and equal month weights. Within each seed, bold marks the lowest raw and clipped TEST energy scores. Raw coverage is the percentage of observations inside 90\% prediction intervals, averaged over five quantities and twelve hidden positions. Each fit supplies its own 39 ensemble members.}
\label{tab:seeds}
\begin{tabular}{lcR{00000}R{00000}R{0.0000}R{0.0000}c}
\toprule
 & & \multicolumn{2}{c}{\shortstack{Step at lowest\\validation score}} & \multicolumn{2}{c}{Energy score} & Raw coverage \\
\cmidrule(lr){3-4}\cmidrule(lr){5-6}
Method & Seed & \multicolumn{1}{c}{Raw} & \multicolumn{1}{c}{Clipped} & \multicolumn{1}{c}{Raw} & \multicolumn{1}{c}{Clipped} & (\%) \\
\midrule
D & 7 & 9560 & 29636 & 1.9575 & 1.9147 & 67.2 \\
D & 13 & 9560 & 9560 & 1.8297 & 1.8213 & 75.9 \\
D & 42 & 16252 & 16252 & 1.8758 & 1.8724 & 75.4 \\ \addlinespace[2pt]
I-S & 7 & 9560 & 9560 & 1.9423 & 1.9385 & 63.3 \\
I-S & 13 & 22944 & 22944 & 1.8954 & 1.8826 & 66.5 \\
I-S & 42 & 16252 & 16252 & 1.8070 & 1.8071 & 78.2 \\ \addlinespace[2pt]
O-S & 7 & 29636 & 29636 & 1.9558 & 1.9116 & 60.4 \\
O-S & 13 & 29636 & 29636 & 1.8934 & 1.8824 & 70.5 \\
O-S & 42 & 16252 & 16252 & \textbf{1.8056} & \textbf{1.8045} & 79.2 \\ \addlinespace[2pt]
G5 & 7 & 29636 & 29636 & \textbf{1.7236} & \textbf{1.7239} & 75.9 \\
G5 & 13 & 29636 & 29636 & \textbf{1.7853} & \textbf{1.7889} & 67.1 \\
G5 & 42 & 16252 & 16252 & 1.8453 & 1.8404 & 84.0 \\
\bottomrule
\end{tabular}
\end{table}

\begin{table}[ht]
\centering
\caption{Primary energy score differences, first method minus second. Negative values favor the first method. Unmarked forecasts are raw. Superscript $c$ marks clipped forecasts at the checkpoint with the lowest clipped validation score. The first four rows assess O-S superiority. Both methods receive the same resampled years or groups of two consecutive years, with their original fits and ensembles.}
\label{tab:contrasts}
\setlength{\tabcolsep}{2pt}
\setlength{\cmidrulekern}{2pt}
\small
\begin{tabular}{@{}p{81pt}>{$}r<{$}!{\extracolsep{2.1pt}}ccrrr@{}}
\toprule
& & \multicolumn{2}{c}{Temporal 95\% interval} & \multicolumn{3}{c@{}}{Contrast by seed} \\
\cmidrule(lr){3-4}\cmidrule(l){5-7}
Contrast & \multicolumn{1}{c}{Mean} & Year & 2 years & \multicolumn{1}{c}{7} & \multicolumn{1}{c}{13} & \multicolumn{1}{c@{}}{42} \\
\midrule
O-S minus D & -0.0028 & $[-0.0346,0.0242]$ & $[-0.0345,0.0233]$ & $-0.0018$ & $0.0637$ & $-0.0702$ \\
O-S minus D$^c$ & 0.0154 & $[-0.0024,0.0312]$ & $[-0.0034,0.0320]$ & $0.0410$ & $0.0721$ & $-0.0669$ \\
O-S minus G5 & 0.1002 & $[0.0712,0.1391]$ & $[0.0671,0.1422]$ & $0.2322$ & $0.1080$ & $-0.0398$ \\
O-S minus G5$^c$ & 0.1005 & $[0.0716,0.1396]$ & $[0.0677,0.1420]$ & $0.2319$ & $0.1045$ & $-0.0348$ \\
\addlinespace[2pt]
O-S minus I-S & 0.0033 & $[-0.0236,0.0349]$ & $[-0.0259,0.0431]$ & $0.0135$ & $-0.0020$ & $-0.0015$ \\
I-S minus D & -0.0061 & $[-0.0562,0.0330]$ & $[-0.0720,0.0410]$ & $-0.0152$ & $0.0656$ & $-0.0688$ \\
O-S minus\newline Gaussian interpolation & -0.2402 & $[-0.5298,-0.0481]$ & $[-0.4539,-0.0882]$ & $-0.1694$ & $-0.2318$ & $-0.3196$ \\
\bottomrule
\end{tabular}
\end{table}

\FloatBarrier
\subsection{Components of the energy score}

The energy score subtracts half the mean distance between distinct ensemble members from their mean distance to the observation. Figure~\ref{fig:calibration}a separates these contributions for G5 relative to D. Larger distances between its members account for most of G5's mean score advantage. Its members are also slightly closer to the observations on average.

At seed 13, G5 reduces both mean distances. The decrease in distance to the observation outweighs half the decrease in distance between members, lowering the score. At seed 42, both mean distances increase. Half the increase in distance between members exceeds the increase in distance to the observation, again lowering the score. G5 changes representation and training loss in the same experiment. Either change could contribute to its lower score.

\FloatBarrier
\subsection{Where prediction intervals miss}

Similar energy scores can conceal different errors in the prediction intervals. O-S falls 2.8 percentage points farther below 90\% coverage than D, exceeding the threshold of two percentage points (Table~\ref{tab:calibration}). It covers the variance of field magnitude more often than D, but every other primary quantity less often (Table~\ref{tab:channelcoverage}). Raw Gaussian interpolation comes closer to 90\% coverage than either D or O-S, yet its 50\% intervals contain 76.4\% of observations. These fractions count individual coordinates across TEST windows. Conditional coverage concerns windows with a given set of available measurements. Simultaneous coverage asks how often all 60 intervals contain their observations in the same window.

G5's low coverage is consistent with intervals that are too narrow, poorly centered, or both. The fractions below the lower endpoint and above the upper endpoint show which side accounts for more misses. The 90\% interval runs from the second to the 38th of the 39 ordered members and includes both endpoints. Independent sampling of the members and observation from the same continuous conditional law gives expected fractions of 5\%, 90\%, and 5\%. These rank probabilities also hold under exchangeability when all 40 values are distinct \citep{gneiting2007calibration}. Observations equal to an interval endpoint count toward coverage and can change the fractions.

\begin{figure}[!htbp]
 \centering
 \includegraphics[width=\linewidth]{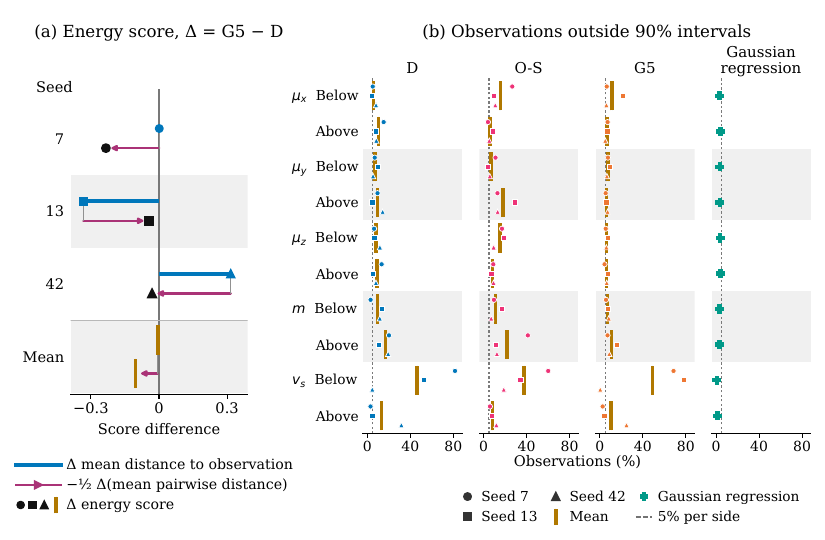}
\caption{Each diffusion fit uses its checkpoint with the lowest raw validation energy score. (a) G5 minus D at each seed and across the three fits. Starting at zero, the blue line ends at the change in mean distance from members to the observation. From there the arrow subtracts half the change in mean distance between distinct ensemble members. The arrow ends at the change in energy score on 60 coordinates standardized on TRAIN. (b) Percentages of observations below and above their nominal 90\% prediction intervals for each quantity and method. In both plots, gold bars show the mean across the three seeds for each diffusion result. Crosses show Gaussian regression, with one ensemble per window. The dashed line marks the expected 5\% on each side of the interval when the observation and the 39 members are exchangeable and all 40 values are distinct.}
 \label{fig:calibration}
\end{figure}

For each fit, the fractions below and above the intervals average the twelve hidden positions in each TEST window, then windows within each month, and then months with equal weight. Each of the three diffusion fits receives equal weight in the method mean.

For G5, 49.0\% of the withheld $v_s$ values fall below the intervals and 10.5\% above, averaged across the three fits. For O-S, the corresponding fractions are 37.7\% and 8.4\%. For both G5 and D, most misses for $v_s$ fall below the intervals at seeds 7 and 13 and above them at seed 42 (Figure~\ref{fig:calibration}b).

\FloatBarrier

For each of the six stochastic methods, both resampling schemes estimate the fractions below and above the intervals separately for each primary quantity. Along with the four methods in Figure~\ref{fig:calibration}b, the resampling includes I-S and Gaussian interpolation. For G5's $v_s$ intervals, subtracting the fraction above from the fraction below before rounding gives 38.4 percentage points. The marginal 95\% interval is $[36.2,40.6]$ percentage points when both fractions use the same 10{,}000 resamples of individual years, including repeated years. It describes variation across years for the original fits, training steps, and ensembles. Repeating training and validation would add further variation.

Gaussian regression's wide intervals contain 98.1\% of $v_s$ observations, with 0.6\% below and 1.2\% above. This exceeds the intended 90\%, even though its primary energy score is worse than G5's. Width helps explain how a forecast can cover more observations and still score worse.

\begin{table}[ht]
\centering
\caption{Marginal forecast accuracy and interval coverage for the five primary quantities. Standardization using the training data makes the continuous ranked probability score and interval width dimensionless. Coverage is the percentage of observations inside the interval. Values average hidden positions and windows within months, give months equal weight, and then average the three diffusion fits. Other methods contribute one ensemble per window. Raw and clipped diffusion forecasts use the checkpoints with the lowest validation energy scores for each form. The extra D row clips samples from the raw forecast's checkpoint. The lowest continuous ranked probability score is in bold.}
\label{tab:calibration}
\setlength{\tabcolsep}{4pt}
\begin{tabular}{lcR{00.0}cR{00.0}c}
\toprule
 & & \multicolumn{2}{c}{50\% intervals} & \multicolumn{2}{c}{90\% intervals} \\
\cmidrule(lr){3-4}\cmidrule(lr){5-6}
Procedure & \parbox[c]{27mm}{\centering Continuous ranked probability score} & \multicolumn{1}{c}{\parbox[c]{13.5mm}{\centering Coverage (\%)}} & Width & \multicolumn{1}{c}{\parbox[c]{13.5mm}{\centering Coverage (\%)}} & Width \\
\midrule
D raw & 0.1753 & 35.5 & 0.2878 & 72.8 & 0.8709 \\
D clipped & 0.1721 & 33.5 & 0.2814 & 71.0 & 0.8617 \\
D clipped, raw validation step & 0.1734 & 36.1 & 0.2883 & 73.5 & 0.8722 \\
I-S raw & 0.1745 & 32.7 & 0.2785 & 69.4 & 0.8315 \\
I-S clipped & 0.1735 & 32.8 & 0.2789 & 69.6 & 0.8319 \\
O-S raw & 0.1729 & 31.9 & 0.2691 & 70.0 & 0.8213 \\
O-S clipped & 0.1701 & 32.6 & 0.2685 & 71.1 & 0.8195 \\
G5 raw & 0.1634 & 38.1 & 0.3137 & 75.7 & 0.9308 \\
G5 clipped & \textbf{0.1624} & 38.4 & 0.3128 & 75.8 & 0.9287 \\
Gaussian interpolation raw & 0.1960 & 76.4 & 0.6262 & 94.3 & 1.5822 \\
Gaussian interpolation clipped & 0.1935 & 68.6 & 0.5512 & 93.6 & 1.3933 \\
Gaussian regression raw & 0.1813 & 75.6 & 0.5617 & 94.0 & 1.4203 \\
Gaussian regression clipped & 0.1804 & 68.8 & 0.5103 & 93.5 & 1.2863 \\
Linear interpolation raw & 0.1993 & 0.0 & 0.0000 & 0.0 & 0.0000 \\
Linear interpolation clipped & 0.1993 & 0.0 & 0.0000 & 0.0 & 0.0000 \\
Persistence raw & 0.2697 & 0.1 & 0.0000 & 0.1 & 0.0000 \\
Persistence clipped & 0.2697 & 0.1 & 0.0000 & 0.1 & 0.0000 \\
\bottomrule
\end{tabular}
\end{table}

\begin{table}[ht]
\centering
\caption{Percentage of observations inside each primary quantity's 90\% prediction intervals before clipping. Each fit's coverage averages hidden positions and windows within months, then gives months equal weight. Diffusion results average the three fits.}
\label{tab:channelcoverage}
\begin{tabular}{lrrrrr}
\toprule
Procedure & \multicolumn{1}{c}{$\mu_x$} & \multicolumn{1}{c}{$\mu_y$} & \multicolumn{1}{c}{$\mu_z$} & \multicolumn{1}{c}{$m$} & \multicolumn{1}{c}{$v_s$} \\
\midrule
D & 83.6 & 83.2 & 83.0 & 73.8 & 40.6 \\
I-S & 82.2 & 74.8 & 80.7 & 74.3 & 34.9 \\
O-S & 78.2 & 74.6 & 76.4 & 67.0 & 53.9 \\
G5 & 82.1 & 85.9 & 87.6 & 82.2 & 40.5 \\
Gaussian interpolation & 93.1 & 93.5 & 92.2 & 94.1 & 98.5 \\
Gaussian regression & 92.8 & 92.9 & 92.2 & 93.9 & 98.1 \\
Linear interpolation & 0.0 & 0.0 & 0.0 & 0.0 & 0.0 \\
Persistence & 0.1 & 0.2 & 0.1 & 0.0 & 0.0 \\
\bottomrule
\end{tabular}
\end{table}

\FloatBarrier
\subsection{Other magnetic quantities}

Which forecast scores best depends on which magnetic quantities enter the score (Table~\ref{tab:alternative}). The three vector components give 36 standardized coordinates over the missing hour. Including all six quantities gives 72 and puts additional weight on vector variance, which the other five determine for realizable moments. The sixth coordinate enters both the distance to the withheld record and the distance between ensemble members. Within each member, the equality residual compares its $v_v$ with the value implied by its other five predictions.

Raw Gaussian regression and G5 score similarly on the vector means alone. Their scores differ more on all six quantities, even though both methods derive vector variance from the same equality. Among raw forecasts, G5 leads on the five primary quantities and O-S on all six. For linear interpolation on TEST, clipping changes only $v_v$, leaving the primary score unchanged and worsening the score on all six quantities.

Hourly magnetic energy poses a different question. For realizable moments, $m^2+v_s$ is the mean squared strength of the five field vectors in a bin, each averaged over one minute. After conversion to T$^2$ and division by $2\mu_0$, averaging across the twelve bins gives one energy density per member. This quantity omits variation within the original minutes. It is consistent with all six predictions only if five vectors in each bin can have those moments.

CRPS compares these scalar predictions with the same transformation of the withheld record. For example, reversing the sign of every mean vector leaves the hourly energy density unchanged. Raw Gaussian regression has the lowest mean CRPS for hourly energy, slightly below raw Gaussian interpolation. Clipping brings the Gaussian regression moments within the physical bounds in Equation~\ref{eq:support} to rounding tolerance. Its 90\% intervals widen on average. The lower endpoints rise by 1.76 picojoules per cubic meter on average, and the percentage of withheld values below their own intervals rises from 2.71\% to 30.83\%. Coverage falls because the wider intervals exclude more observations at the lower end.

\begin{table}[ht]
\centering
\caption{Forecast accuracy for the vector components, all six magnetic quantities, and hourly magnetic energy density. The energy score uses 36 standardized coordinates for the vector components and 72 for all six quantities. Each member's twelve values of $m^2+v_s$ determine its mean magnetic energy density over the missing hour. The calculation uses one vector average per minute and omits variation within each minute. The same calculation applies to raw forecasts, including combinations of moments that five field vectors cannot produce. The continuous ranked probability score and interval width are in $10^{-12}$ J\,m$^{-3}$. Coverage is in percent. Each diffusion fit uses the checkpoint with the lowest validation energy score for the forecast form shown. Bold marks the lowest score in each score column, including ties at the precision shown.}
\label{tab:alternative}
\setlength{\tabcolsep}{2.9pt}
\begin{tabular}{@{}lcccR{00.0}R{00.0000}@{}}
\toprule
 & \multicolumn{2}{c}{Energy score} & \multicolumn{3}{c}{Hourly magnetic energy density} \\
\cmidrule(lr){2-3}\cmidrule(lr){4-6}
Procedure & Vector & \parbox[c]{17mm}{\centering All six quantities} & \parbox[c]{27mm}{\centering Continuous ranked probability score} & \multicolumn{1}{c}{\parbox[c]{19mm}{\centering 90\% coverage (\%)}} & \multicolumn{1}{c@{}}{\parbox[c]{14mm}{\centering 90\% width}} \\
\midrule
D raw & 1.6351 & 2.1528 & 2.8599 & 74.3 & 6.2166 \\
D clipped & 1.6222 & 2.5726 & 2.7568 & 68.5 & 6.6024 \\
I-S raw & 1.6435 & 2.1590 & 2.4664 & 72.4 & 6.6657 \\
I-S clipped & 1.6435 & 2.8791 & 2.3049 & 73.1 & 6.8576 \\
O-S raw & 1.6316 & \textbf{2.1503} & 2.9225 & 65.3 & 6.5977 \\
O-S clipped & 1.6316 & 2.3863 & 2.7461 & 72.2 & 6.6863 \\
G5 raw & 1.5740 & 2.4959 & 1.8961 & 82.2 & 6.8060 \\
G5 clipped & 1.5740 & 2.4285 & 1.9066 & 82.4 & 6.8757 \\
Gaussian interpolation raw & 1.5909 & 2.7342 & 1.0816 & 93.8 & 6.1012 \\
Gaussian interpolation clipped & 1.5909 & 4.2146 & 2.5465 & 64.8 & 11.2941 \\
Gaussian regression raw & \textbf{1.5728} & 6.1746 & \textbf{1.0497} & 94.0 & 6.0961 \\
Gaussian regression clipped & \textbf{1.5728} & 5.0391 & 2.1873 & 67.5 & 10.0897 \\
Linear interpolation raw & 1.9834 & 2.8920 & 1.2771 & 0.0 & 0.0000 \\
Linear interpolation clipped & 1.9834 & 4.6381 & 1.2771 & 0.0 & 0.0000 \\
Persistence raw & 2.7925 & 3.6601 & 2.6284 & 0.0 & 0.0000 \\
Persistence clipped & 2.7925 & 3.6601 & 2.6284 & 0.0 & 0.0000 \\
\bottomrule
\end{tabular}
\end{table}

\FloatBarrier
\subsection{Temporal differences}

A reconstruction can make small errors at the edges of a gap and still vary too little inside it. Table~\ref{tab:geometry} compares mean vector increment magnitudes with the measured record at the two edges and inside the gap. At either edge, the predicted and observed increments share the same available endpoint, so their difference has the same norm as the vector prediction error at the adjacent hidden bin.

The mean magnitude of the measured increments is slightly larger inside the gap than at either edge. Linear interpolation has the lowest root mean square error at both edges, although its mean increment magnitude is smaller than the measured mean there and inside the gap. All four diffusion methods give larger mean magnitudes than observed at both edges. Inside the gap, D, I-S, and O-S give smaller means, while G5 comes closest to the measured mean. Both Gaussian methods give larger means at the edges and inside the gap. Opposite changes can have the same magnitude.

The variogram score compares differences between coordinates, making it sensitive to how quantities vary together \citep{scheuerer2015variogram}. With exponent $1/2$, it gives equal weight to 55 comparisons between consecutive times for the same quantity and 120 between quantities at the same time. For each coordinate pair $(i,j)$, define $d_k=|x_{k,i}-x_{k,j}|^{1/2}$. Its contribution is $(|y_i-y_j|^{1/2}-\bar d)^2-s_d^2/K$, where $K=39$ and $s_d^2$ uses divisor $K-1$. The square compares the observed transformed difference with the ensemble mean. For iid members with finite $\operatorname{Var}(d_k)$, subtracting $s_d^2/K$ removes the upward bias from estimating that mean. Averaging the 175 contributions measures agreement in these differences. Adding the same constant to every standardized forecast coordinate leaves all 175 differences unchanged, even if the forecast moves farther from the observation.

\begin{table}[ht]
\centering
\caption{Vector increments and variogram scores before clipping. Increment magnitude and root mean square error (RMSE) are in nT. Each window's RMSE uses all 39 predicted increments. Its square equals the squared distance from their mean to the observed increment plus their mean squared distance from their own mean. The table averages window RMSEs within months, then across months and diffusion fits. The variogram score averages contributions from 175 differences between standardized coordinates after the adjustment for ensemble size described above. Clipping leaves the vector increments of a given ensemble unchanged but can change the variogram through $m$ and $v_s$. At seed 7, D's raw and clipped forecasts use different checkpoints, so their predicted vectors can differ. Bold marks the lowest variogram score and the lowest RMSE at each edge among the forecasts.}
\label{tab:geometry}
\setlength{\tabcolsep}{4.8pt}
\begin{tabular}{lcccccc}
\toprule
 & & \multicolumn{3}{c}{Mean increment magnitude} & \multicolumn{2}{c}{Root mean square error} \\
\cmidrule(lr){3-5}\cmidrule(lr){6-7}
Procedure & Variogram & Left edge & Right edge & Internal & Left edge & Right edge \\
\midrule
Measured record & --- & 1.0460 & 1.0539 & 1.0831 & --- & --- \\
\addlinespace[2pt]
D & 0.0524 & 1.3121 & 1.3581 & 0.8560 & 1.8164 & 1.8323 \\
I-S & 0.0521 & 1.3309 & 1.4224 & 0.8156 & 1.8059 & 1.8854 \\
O-S & 0.0524 & 1.3111 & 1.3296 & 0.7635 & 1.7974 & 1.7990 \\
G5 & \textbf{0.0490} & 1.3717 & 1.3830 & 1.0789 & 1.8893 & 1.8958 \\
Gaussian interpolation & 0.0905 & 1.5737 & 1.5587 & 1.5718 & 2.1020 & 2.1114 \\
Gaussian regression & 0.0796 & 1.5734 & 1.5647 & 1.5642 & 2.0780 & 2.0770 \\
Linear interpolation & 0.0658 & 0.2518 & 0.2518 & 0.2518 & \textbf{1.0373} & \textbf{1.0309} \\
Persistence & 0.1003 & 0.0000 & 3.2730 & 0.0000 & 1.0460 & 3.1140 \\
\bottomrule
\end{tabular}
\end{table}

\FloatBarrier
\subsection{Numerical precision}
\label{app:numerics}

A predicted bin counts as a violation when any of the three margins $m-r$, $v_s$, or $H-v_s$ is negative by more than its rounding tolerance, where $r=\|\mu\|$ and $H=\frac32m^2+\frac52r^2$. Define $A(a,b;q)=64\epsilon_{64}\max(|a|,|b|,q)$, with $\epsilon_{64}=2^{-52}$. For float64 storage, the corresponding tolerances are $A(m,r;B_{\mathrm{ref}})$, $A(v_s,0;B_{\mathrm{ref}}^2)$, and $A(H,v_s;B_{\mathrm{ref}}^2)$.

For float32 storage, let $b_j$ be the TRAIN scale of coordinate $j$ times half the larger finite adjacent spacing at its nearest float32 standardized value, with ties rounded to even. Put $b_r=(b_{\mu_x}^2+b_{\mu_y}^2+b_{\mu_z}^2)^{1/2}$, $h_m=2|m|b_m+b_m^2$, and $h_r=2rb_r+b_r^2$. Increase the three float64 tolerances by $b_m+b_r$, $b_{v_s}$, and $b_{v_s}+\frac32h_m+\frac52h_r$, respectively. Table~\ref{tab:main} gives the percentage of bins that fail this check alongside the mean absolute equality residual.

Rounding can change the primary coordinates even when algebraic completion gives an equality residual of zero. Table~\ref{tab:roundtrip} compares the original values in 5{,}760 hidden observation bins with those recovered after normalization and inversion. Rounding five or six standardized coordinates to single precision gives the same errors in the five primary coordinates. With five coordinates, the decoder computes $v_v$ after inversion. All 5{,}760 bins satisfy the inequalities in all four representations.

\begin{table}[t]
\centering
\caption{Numerical precision for 5{,}760 hidden observation bins. The first row gives the original measurements. The other three rows convert them to standardized coordinates and back to physical units, with both transformations in double precision. The last two rows round the standardized coordinates to single precision before inversion. Maximum $|\Delta u|$ is the largest change across the five primary coordinates and hidden bins after division by the corresponding training scales, and is dimensionless. Equality residuals are in nT$^2$. Deriving $v_v$ makes that residual zero while the primary coordinates still contain rounding errors.}
\label{tab:roundtrip}
\begin{tabular}{llll}
\toprule
Representation & \multicolumn{1}{c}{\parbox[c]{16mm}{\centering Maximum $|\Delta u|$}} & \multicolumn{1}{c}{Mean $|R_{\rm obs}|$} & \multicolumn{1}{c}{\parbox[c]{16mm}{\centering Maximum $|R_{\rm obs}|$}} \\
\midrule
Original, double precision & \multicolumn{1}{c}{0} & $8.476\times10^{-15}$ & $3.326\times10^{-13}$ \\
Double precision, normalized and inverted & $1.271\times10^{-15}$ & $8.987\times10^{-15}$ & $5.573\times10^{-13}$ \\
Single precision, six coordinates & $4.977\times10^{-7}$ & $1.571\times10^{-6}$ & $6.268\times10^{-5}$ \\
Single precision, five coordinates & $4.977\times10^{-7}$ & \multicolumn{1}{c}{0} & \multicolumn{1}{c}{0} \\
\bottomrule
\end{tabular}
\end{table}

\FloatBarrier
\section{Variation across test cases}
\label{app:strata}

\FloatBarrier
\subsection{Group definitions}

Which differences between test cases could guide a forecast? The spacecraft and date are known before reconstruction. Table~\ref{tab:strata} also divides windows into thirds by mean field magnitude and, separately, by mean vector increment length. Both comparisons use measurements inside the gap. The increment statistic averages the lengths of the eleven changes between consecutive mean field vectors in the twelve hidden bins. Selecting windows by their outcomes changes the distribution against which the forecasts are scored. Even a forecast that gives the correct distribution conditional on the available measurements can change rank within such a group \citep{lerch2017forecaster}. D scores lower than O-S in the lowest magnitude third, and O-S scores lower in the highest. For a new gap, choosing between the methods on this basis would depend on predicting its magnitude group from the observations on either side.

Within a group, each represented month contributes its mean score. Window counts vary across months within a group. Weighting the group means by their numbers of windows can give a different result from weighting months equally across TEST. For each comparison, the marginal 95\% interval comes from resampling the years in that group and reusing the forecasts for each window. In every TEST year, the mean score over the three fits per method is lower for G5 than for O-S (Table~\ref{tab:years}). At seed 42, O-S scores lower than G5 when averaged over the same 20 years (Table~\ref{tab:contrasts}).

Differences by spacecraft are also difficult to separate from changes over time. In 2020--2024, Wind supplied 119 windows and ACE supplied one. The calendar groups cover different phases of the solar cycle.

Each third contains 160 of the 480 TEST windows. Each cutoff falls strictly between consecutive observed values. The two cutoffs are 3.9350 and 5.7891 nT for mean field magnitude and 0.6134 and 1.2055 nT for mean vector increment length.

\begin{table}[ht]
\centering
\caption{Raw primary energy scores by spacecraft and calendar period. Two further comparisons divide test windows into thirds by mean field magnitude or by mean vector increment length during the missing hour. Scores use standardized coordinates and give the months present in each group equal weight. Each diffusion method contributes the same three fits to every group. The final column gives the marginal 95\% interval for O-S minus D from resampling years. Grouping by measurements inside the gap describes performance retrospectively. ACE is the Advanced Composition Explorer. Bold marks the lowest score among D, O-S, and G5 in each row.}
\label{tab:strata}
\setlength{\tabcolsep}{3.75pt}
\begin{tabular}{@{}lR{000}cccM{-0.0000}c@{}}
\toprule
Stratum & \multicolumn{1}{c}{Windows} & D & O-S & G5 & \multicolumn{1}{c}{O-S minus D} & Year interval \\
\midrule
Wind & 335 & 1.7136 & 1.7119 & \textbf{1.6159} & -0.0017 & $[-0.0382,0.0303]$ \\
ACE & 145 & 1.8597 & 1.9008 & \textbf{1.7886} & 0.0411 & $[-0.0009,0.0869]$ \\
\addlinespace[2pt]
2005--2009 & 120 & 1.9817 & 2.0219 & \textbf{1.9051} & 0.0403 & $[0.0158,0.0638]$ \\
2010--2015 & 144 & 1.7735 & 1.7728 & \textbf{1.6890} & -0.0007 & $[-0.0416,0.0415]$ \\
2016--2019 & 96 & 1.6349 & 1.6220 & \textbf{1.5657} & -0.0129 & $[-0.0543,0.0237]$ \\
2020--2024 & 120 & 2.1330 & 2.0928 & \textbf{1.9545} & -0.0403 & $[-0.1384,0.0293]$ \\
\addlinespace[2pt]
Magnitude, lower third & 160 & 1.1777 & 1.2324 & \textbf{1.1010} & 0.0547 & $[0.0195,0.0835]$ \\
Magnitude, middle third & 160 & 1.4606 & 1.4747 & \textbf{1.4339} & 0.0140 & $[-0.0044,0.0310]$ \\
Magnitude, upper third & 160 & 3.3910 & 3.3366 & \textbf{3.1849} & -0.0544 & $[-0.1089,-0.0023]$ \\
\addlinespace[2pt]
Increment, lower third & 160 & 0.8924 & 0.9174 & \textbf{0.8255} & 0.0250 & $[-0.0056,0.0560]$ \\
Increment, middle third & 160 & 1.4411 & 1.4416 & \textbf{1.3768} & 0.0005 & $[-0.0392,0.0434]$ \\
Increment, upper third & 160 & 3.2522 & 3.2284 & \textbf{3.0924} & -0.0238 & $[-0.0804,0.0271]$ \\
\bottomrule
\end{tabular}
\end{table}

\begin{table}[ht]
\centering
\caption{The upper table gives annual raw primary energy scores in standardized coordinates for 24 TEST windows per year. Scores give equal weight to windows within each month, to the two months within each fit, and to the three fits of each diffusion method. Gaussian interpolation contributes one ensemble per window. Bold marks the lowest score among the five methods in each year. The lower table counts the years in which the first method scores lower or higher than the second. Each annual difference subtracts the second method's score from the first, so a negative median favors the first method.}
\label{tab:years}
\begin{tabular}{lccccc}
\toprule
Year & D & I-S & O-S & G5 & \shortstack{Gaussian\\interpolation} \\
\midrule
2005 & 3.3402 & 3.4233 & 3.3349 & \textbf{3.1962} & 3.3435 \\
2006 & \textbf{1.8130} & 1.8449 & 1.8911 & 1.8222 & 2.0005 \\
2007 & 1.5185 & 1.6045 & 1.5749 & \textbf{1.4551} & 2.1359 \\
2008 & 1.8584 & 1.8789 & 1.8882 & \textbf{1.7878} & 2.1347 \\
2009 & 1.3782 & 1.4624 & 1.4206 & \textbf{1.2640} & 1.5893 \\
2010 & 1.2774 & 1.3131 & 1.2785 & \textbf{1.2081} & 1.5417 \\
2011 & 2.5543 & 2.5698 & 2.5144 & 2.3807 & \textbf{2.2021} \\
2012 & 1.3628 & 1.3747 & 1.4155 & \textbf{1.3091} & 1.6150 \\
2013 & 1.1809 & 1.2399 & 1.2585 & \textbf{1.1578} & 1.4643 \\
2014 & 2.5162 & \textbf{2.4384} & 2.4978 & 2.4404 & 2.4855 \\
2015 & 1.7493 & 1.7088 & 1.6719 & \textbf{1.6383} & 1.7181 \\
2016 & 1.9325 & 1.8200 & 1.8618 & \textbf{1.7919} & 1.9331 \\
2017 & 1.7522 & 1.7971 & 1.7414 & \textbf{1.7083} & 2.0006 \\
2018 & 1.5171 & 1.5611 & 1.5522 & \textbf{1.4413} & 1.6696 \\
2019 & 1.3378 & 1.3602 & 1.3326 & \textbf{1.3214} & 1.6536 \\
2020 & 1.5016 & 1.5137 & 1.5409 & \textbf{1.4375} & 1.6264 \\
2021 & 1.3806 & 1.4460 & 1.4136 & \textbf{1.3686} & 1.5898 \\
2022 & 1.9837 & 2.0144 & 1.9358 & \textbf{1.9206} & 1.9962 \\
2023 & 2.5198 & 2.3208 & 2.5221 & \textbf{2.1314} & 2.1605 \\
2024 & 3.2793 & 2.9392 & 3.0514 & \textbf{2.9142} & 5.6425 \\
\bottomrule
\end{tabular}
\par\medskip
\begin{tabular}{lR{00}R{00}M{-0.0000}}
\toprule
Contrast & \multicolumn{1}{c}{First lower (years)} & \multicolumn{1}{c}{First higher (years)} & \multicolumn{1}{c}{Median difference} \\
\midrule
O-S minus D & 9 & 11 & 0.0017 \\
G5 minus D & 19 & 1 & -0.0700 \\
G5 minus O-S & 20 & 0 & -0.1006 \\
\bottomrule
\end{tabular}
\end{table}

\FloatBarrier
\subsection{Cancellation groups}

When field vectors point in different directions, their average can be small even at constant field strength. How does the difference in energy score between O-S and D vary with cancellation in the available measurements? One grouping of TEST windows uses 18 bins before the gap and 18 after it. All 36 bins span five minutes each and are available at prediction time. The second grouping uses measurements within the missing hour. For each bin, $C=(m^2-\|\mu\|^2)/m^2$ measures cancellation relative to its squared mean magnitude ($C=0$ when $m=0$). The measure $A=(m^2-\|\mu\|^2)/B_{\mathrm{ref}}^2$ divides the same difference by the mean of $m^2+v_s$ across distinct training bins. Both measures are dimensionless.

The first rule averages $C$ over the 36 available bins. The second averages $A$ over the twelve hidden bins. The TRAIN 75th percentiles are 0.044395 for mean $C$ in the available bins and 0.022696 for mean $A$ in the missing hour. A window is in the high group when its mean exceeds the corresponding cutoff. These thresholds weight all 122{,}309 overlapping TRAIN windows equally, whereas normalization uses distinct bins. The analysis followed the previously documented grouping rules and was run after inspection of the primary results. These comparisons are exploratory.

Across all twenty TEST years, 128 windows have high cancellation in the available measurements. O-S scores lower than D at all three seeds. The mean difference, O-S minus D, is $-0.0256$, with a marginal 95\% interval of $[-0.0740,0.0147]$ from resampling years. G5 scores lower than O-S in this group, with an O-S minus G5 difference of $0.0573$ and a marginal 95\% interval of $[0.0053,0.1246]$. In the other 352 windows, D scores lower than O-S, with an O-S minus D difference of $0.0114$.

For O-S minus D, subtracting the mean in the other windows from the mean in windows with high cancellation gives $-0.0370$, with a marginal 95\% interval of $[-0.0778,0.0012]$. For each of the 10{,}000 bootstrap replicates, both groups use the same sample of twenty TEST years taken with replacement. Each occurrence of a year contributes the months present in that group, with equal weight per month. Every resample uses the same fits and ensemble members. The interval for this exploratory contrast includes zero, so the difference between groups is uncertain.

Among the 121 windows with high cancellation in the missing hour, O-S minus D is $-0.0128$, with a marginal 95\% interval of $[-0.0858,0.0629]$. The difference is $0.0068$ in the other 359 windows. TRAIN supplies the thresholds, but the groups depend on measurements inside each TEST gap, so these scores assess performance after outcome selection.

Energy groups use window means of $Q/B_{\mathrm{ref}}^2$, where $Q=m^2+v_s$ is the mean squared field strength in nT$^2$. The ratio is dimensionless and concerns a physical quantity distinct from the statistical energy score. The TRAIN energy quartile cutoffs are 0.349809, 0.571606, and 1.078729 for the available bins and 0.343518, 0.563106, and 1.069013 for the missing hour. The companion lists the TEST windows in each group. A window on an energy cutoff enters the higher quartile. A window on the cancellation cutoff stays in the rest group.

For each cancellation rule, high cancellation and its complement form two broad groups; crossing them with energy quartiles forms eight more. Adding the full TEST set to the ten groups from each rule brings the total to 21. Each compares D, I-S, O-S, and G5 in three forms: before clipping, after clipping at the step with the lowest raw validation score, and after clipping at the step with the lowest clipped validation score. Raw and clipped versions of Gaussian interpolation, linear interpolation, and persistence bring the total to 18 procedures, excluding Gaussian regression. Scores give the represented months equal weight. Each diffusion score also averages over three fits.

Among windows with high cancellation during the missing hour, seven from five years fall in the lowest energy quartile. Identifying these windows required the measurements withheld from the models.

\FloatBarrier
\section{Clipping}
\label{app:extension}

\FloatBarrier
\subsection{Completion and clipping}

Clipping changes a forecast after the model has generated it. To isolate that effect, Table~\ref{tab:operators} scores the same ensemble members before and after clipping. For both evaluations, each diffusion fit uses the checkpoint with the lowest raw validation energy score. Gaussian regression keeps the ridge coefficient and covariance shrinkage with the lowest raw validation energy score. Completing $v_v$ alone removes the equality residual without changing the primary score. Clipping can also move $m$ and $v_s$, two of the five scored quantities.

Clipping retains the mean vector $\mu$. If the predicted mean magnitude $m$ is below $r=\|\mu\|$, clipping first raises it to $r$. It then clips $v_s$ to $[0,1.5m^2+2.5r^2]$ using this value of $m$ and computes $v_v=m^2+v_s-r^2$. The resulting moments satisfy Equation~\ref{eq:support}, so a second application leaves them unchanged to numerical tolerance. Clipping in this order can move a sample farther than projection onto the nearest realizable point.

Let $C$ clip each member of an ensemble $X=(x_1,\ldots,x_K)$ and let $y$ contain the measurements from the missing hour in the same 60 standardized coordinates. The absolute change in energy score is at most twice the mean distance moved,
\begin{equation}
 \left|\operatorname{ES}(CX,y)-\operatorname{ES}(X,y)\right|
 \le \frac{2}{K}\sum_{i=1}^K\|Cx_i-x_i\|.
 \label{eq:movement}
\end{equation}
Moving an ensemble member changes its distance to the observation by at most the distance it moves. For two members, the change in their separation is at most the sum of their displacements. Averaging this bound over distinct pairs shows that half the mean distance between members changes by at most the mean displacement. Equation~\ref{eq:movement} follows by adding the bounds. Either term can increase or decrease. Table~\ref{tab:operators} gives their signed contributions.

\begin{table}[ht]
\centering
\caption{Effect of clipping the same ensemble in 60 primary coordinates standardized using the training data, with equal month weights. Diffusion rows average three fits, each at its checkpoint with the lowest raw validation energy score. Each Gaussian baseline supplies one ensemble per window. Displacement is the mean Euclidean distance an ensemble member moves under clipping. Let $a_K$ be the mean distance to the observation and $b_K$ the mean distance between distinct ensemble members. Here $\Delta$ means clipped minus raw. Subtracting half the change in $b_K$ from the change in $a_K$ gives the change in energy score before rounding.}
\label{tab:operators}
\begin{tabular}{lcM{-0.00000}M{-0.00000}M{-0.00000}}
\toprule Procedure & Displacement & \multicolumn{1}{c}{$\Delta a_K$} & \multicolumn{1}{c}{$-\Delta b_K/2$} & \multicolumn{1}{c}{$\Delta$ energy score} \\
\midrule
D & 0.1254 & -0.00350 & -0.00287 & -0.00637 \\
I-S & 0.0997 & -0.00427 & -0.00119 & -0.00547 \\
O-S & 0.1521 & -0.01952 & 0.00080 & -0.01873 \\
G5 & 0.1055 & 0.00003 & -0.00038 & -0.00035 \\
Gaussian interpolation & 2.1403 & -0.32056 & 0.42478 & 0.10422 \\
Gaussian regression & 1.7733 & -0.15789 & 0.25776 & 0.09987 \\
\bottomrule\end{tabular}
\end{table}
\FloatBarrier

G5 often violates a moment bound (Table~\ref{tab:main}), yet clipping moves its samples only a short distance (Table~\ref{tab:operators}). Clipping moves Gaussian interpolation samples about twenty times as far as G5 samples on average. To measure how far the raw samples fall below the two lower bounds, consider the shortfalls $\max(r-m,0)/B_{\mathrm{ref}}$ and $\max(-v_s,0)/B_{\mathrm{ref}}^2$ before any rounding allowance. Each window contributes the 90th percentile across 468 sample bins (39 ensemble members at twelve hidden positions). The window percentiles are averaged within each month, with equal weight for each monthly mean. The G5 result also averages over the three training seeds.

For the shortfall in mean magnitude, the window 90th percentiles average 0.01648 for G5 and 0.34407 for Gaussian interpolation. For negative variance, the corresponding averages are 0.000342 and 0.017372. The upper variance bound calls for a count because isolated violations can disappear even at the 90th percentile. Across all TEST windows, 5 of 673{,}920 raw G5 sample bins and 0 of 224{,}640 raw Gaussian interpolation sample bins exceed the upper variance bound beyond rounding.

\FloatBarrier
\begin{figure}[ht]
 \centering
 \includegraphics[width=\linewidth]{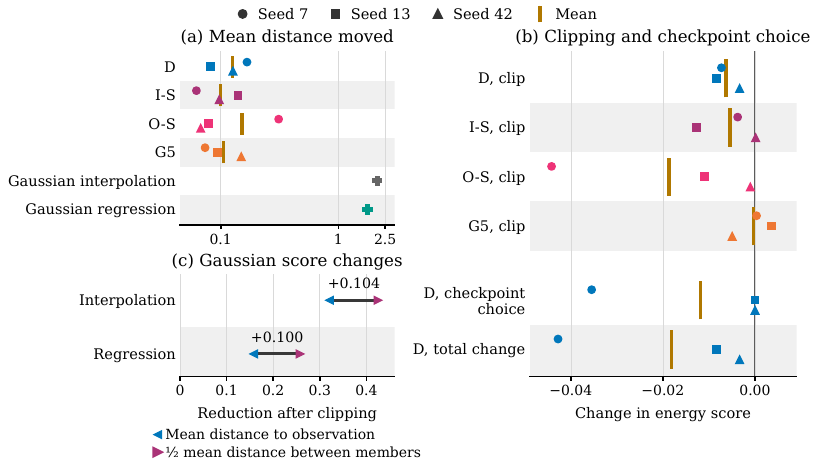}
\caption{Effect of clipping forecast ensembles. Circles, squares, and upward-pointing triangles show results for diffusion seeds 7, 13, and 42, respectively. Gold vertical bars give the mean over three fits. (a) Mean distance an ensemble member moves under clipping, on a logarithmic scale. The two crosses give the mean displacements for Gaussian interpolation and Gaussian regression. (b) The first four rows give the change in TEST energy score after clipping, measured at the training step with the lowest raw validation score for each fit. For D, choosing the checkpoint with the lowest clipped validation score changes the step only at seed 7. The fifth row compares the two checkpoints after clipping. The last row combines that change with clipping at the original checkpoint. (c) For each Gaussian baseline, the left-pointing triangle gives the reduction in mean distance to the observation. The right-pointing triangle gives the reduction in half the mean distance between distinct ensemble members. The horizontal distance between the markers gives the increase in energy score. Scores use 60 coordinates standardized on TRAIN and give months equal weight.}
 \label{fig:operator}
\end{figure}

For D, the raw and clipped scores in Table~\ref{tab:main} reflect two changes. Clipping the original ensemble lowers the mean energy score (Table~\ref{tab:operators}). For D at seed~7, the clipped validation score selects a later checkpoint than the raw validation score (Table~\ref{tab:seeds}). Seeds 13 and 42 keep their original checkpoints. The switch lowers the score by 0.03550 at seed~7. Across the three fits, choosing checkpoints by the clipped validation score lowers the mean TEST score by 0.01183, compared with 0.00637 from clipping at the original checkpoints. Figure~\ref{fig:operator}(b) shows both contributions and their sum for each seed.

\FloatBarrier
\subsection{Gaussian forecasts}

For both Gaussian baselines, clipping reduces mean distance to the measurements in the gap but worsens the primary energy score. The reduction in half the mean distance between members exceeds the reduction in mean distance to the observation. In Figure~\ref{fig:operator}(c), the separation between the two triangles gives the resulting increase in energy score.

Clipping preserves the vector components, so Gaussian regression has the same vector energy score before and after clipping. The same operation changes the predicted mean and variance of field magnitude. Both enter the hourly magnetic energy density through $m^2+v_s$, and the CRPS for that quantity more than doubles (Table~\ref{tab:alternative}).

Before clipping, the difference in mean energy score (G5 minus Gaussian regression) is $-0.1503$. Its marginal 95\% interval lies below zero whether the resampling unit is a year ($[-0.2302,-0.0829]$) or a pair of consecutive years ($[-0.2105,-0.0913]$). The corresponding difference for O-S is $-0.0502$. Its interval includes zero for individual years ($[-0.1336,0.0327]$) and lies below zero for pairs of years ($[-0.0913,-0.0080]$).

Each resampling scheme uses 10{,}000 replicates. Within each replicate, all methods are compared on the same sampled years. Each resample retains the three diffusion fits, the Gaussian ensemble for each window, and the validation choices. The intervals describe variation over the observed test years. These test years were examined before selection of the model configuration.

\FloatBarrier
\section{Penalty strength}
\label{app:fixedfinal}

The averaging penalty acts on an equality, while a realizable forecast must also satisfy the moment bounds. Before this TEST comparison, the study specified three O-S penalty strengths and 29{,}636 training steps for every fit. Table~\ref{tab:fixedfinal} compares D and O-S on forecast accuracy and moment consistency. Each fit generates its ensembles on the same windows with 50 DDIM evaluations per member. Computing the residual penalty adds work to each training step.

\begin{table}[ht]
\centering
 \caption{Forecast quality and moment consistency after 29{,}636 training steps for D ($c_R=0$) and O-S ($c_R>0$). Each row is one fit with equal month weights. The raw and clipped energy scores compare the ensembles with the observations in the five standardized primary quantities at all twelve hidden positions. The raw columns give the mean absolute equality residual $|R_{\mathrm{obs}}|$ in nT$^2$ and the percentage of sample bins in which $m$ or $v_s$ violates a bound in Equation~\ref{eq:support} beyond rounding tolerance. The last two columns give coverage and mean width for the raw 90\% prediction intervals, with width in standardized units. All twelve combinations of coefficient and seed appear at that step, regardless of their test scores. For each seed, bold marks the lowest observed energy score in each score column across the four coefficients.}
\label{tab:fixedfinal}
\begin{tabular}{@{}lccccccc@{}}
\toprule
& & \multicolumn{2}{c}{Energy score} & \multicolumn{4}{c}{Raw samples} \\
\cmidrule(lr){3-4}\cmidrule(l){5-8}
$c_R$ & Seed & Raw & Clipped & $|R|$ & \shortstack{Outside\\range (\%)} & Coverage (\%) & Width \\
\midrule
0 & 7 & 1.9787 & 1.9147 & 3.5152 & 78.4 & 57.5 & 0.8018 \\
 & 13 & 1.9778 & 1.9515 & 3.1092 & 68.1 & 67.9 & 0.8369 \\
 & 42 & 1.9900 & 1.9847 & 4.2064 & 11.4 & 60.6 & 0.7955 \\
\addlinespace[2pt]
0.00005 & 7 & 2.0412 & 1.9714 & 3.5544 & 80.4 & 57.4 & 0.7830 \\
 & 13 & \textbf{1.8735} & \textbf{1.8653} & 2.8988 & 59.7 & 69.4 & 0.8347 \\
 & 42 & 1.9757 & 1.9702 & 3.8648 & 15.5 & 62.9 & 0.8396 \\
\addlinespace[2pt]
0.0005 & 7 & \textbf{1.9558} & \textbf{1.9116} & 2.8245 & 88.8 & 60.4 & 0.7797 \\
 & 13 & 1.8934 & 1.8824 & 2.3812 & 32.5 & 70.5 & 0.8112 \\
 & 42 & 1.9312 & 1.9284 & 3.1843 & 15.6 & 66.8 & 0.8634 \\
\addlinespace[2pt]
0.005 & 7 & 2.0224 & 1.9934 & 1.5068 & 69.1 & 57.2 & 0.7742 \\
 & 13 & 2.0201 & 2.0114 & 1.7024 & 54.4 & 61.4 & 0.8303 \\
 & 42 & \textbf{1.8119} & \textbf{1.8036} & 1.1243 & 44.2 & 68.6 & 0.7964 \\
\bottomrule\end{tabular}
\end{table}

Increasing $c_R$ from 0.0005 to 0.005 lowers the mean absolute equality residual at all three seeds. Each fit contributes 39 ensemble members, each predicting twelve hidden bins per window. The fraction of raw sample bins outside the moment bounds beyond rounding tolerance rises at seeds 13 and 42. At seed 42 it rises from 15.6\% to 44.2\% (Table~\ref{tab:fixedfinal}), even though that fit has the lowest raw energy score of the four coefficients at that seed.

\FloatBarrier
\subsection{Penalty weighting}

Within each fit, $c_R$ stays constant. The Min-SNR data weight $w_t$ varies with noise level, as does the residual coefficient $\lambda_t=c_R/(2\Sigma_t)$ when $c_R>0$. For $c_R=0.0005$, the ratio $\lambda_t/w_t$ is 0.1239, 0.004183, 0.008642, and 1030.7 at $t=1,25,50,100$, respectively. At $t=100$, the Min-SNR data weight has fallen to $2.429\times10^{-7}$. For a batch at one noise level, the gradient of each weighted loss is its coefficient times the gradient of the unweighted loss with respect to the network parameters.

The gradient comparison evaluates both weighted terms on the same batches at the same network parameters, before clipping the total gradient. For each seed, the first and final O-S checkpoints use the same eight TRAIN windows and eight VAL windows, with the same noise draws and $c_R=0.0005$. Each gradient is taken from the mean loss over the eight windows in its batch. At $t\in\{1,25,50,100\}$, the probe evaluates $R_{\mathrm{inst}}$ and $R_{\mathrm{obs}}$ at the denoising mean estimate (M) and the DDIM sample estimate (S). The angle between two nonzero gradients distinguishes reinforcement from opposition.

For S with $R_{\mathrm{obs}}$ at the final O-S checkpoints, the ratio of the residual gradient norm to the data gradient norm ranges from 315 to 1799 on TRAIN and from 214 to 1266 on VAL at $t=100$. At $t=25$, the ratios range from 0.0081 to 0.0267 on TRAIN and from 0.0103 to 0.0465 on VAL. On TRAIN, each fit's residual gradient at $t=100$ has less than one hundredth of its norm at $t=1$. The large ratio at high noise reflects a small data gradient. At $t=100$, the two gradients are nearly perpendicular, with angles of 86.7 to 92.1 degrees on TRAIN and 86.7 to 95.2 degrees on VAL. At each final O-S checkpoint, the gradient of the combined loss at $t=100$ lies within $0.3^\circ$ of the residual gradient on both the TRAIN and VAL batches. Training combines noise levels and clips the total gradient. These batch gradients describe the penalty at two checkpoints. Its accumulated effect also depends on the intervening updates.

\FloatBarrier
\section{Related work}

A model can satisfy an equality at every prediction while assigning the wrong probabilities to the feasible states. In the unit circle experiment of \citet[Section 3.3 and Appendix A.6.1]{bastek2025pidm}, predicting a single point on the circle for every input minimizes the residual loss. A residual can also favor particular states when the objective includes a data loss. In the experiment of \citet[Section~3, Figure~3, and Appendix~F.1.1]{bien2026bias}, the data lie on a circle, while the residual vanishes on the whole unit hypersphere. The authors vary the residual's sensitivity along the circle. For small isotropic perturbations of equal variance, lower sensitivity gives a smaller expected squared residual to leading order. In the three-body experiment of \citet[Section 4.3, Table 5]{zhou2025priors}, trajectory and velocity errors were lower when the model penalized errors in inverse distances and squared velocity components than when it penalized changes in total energy. For the magnetic equality, Equation~\ref{eq:conditional} gives the discrepancy at the conditional mean in terms of conditional variances.

\citet[Section 3.3 and Appendix A.1]{bastek2025pidm} evaluate the residual either at the denoising mean or on a sample approximated with two DDIM steps. Even with the exact gradient of the log density at each noise level, discretizing the reverse process can bias the distribution of samples. The two-state HMM separates that sampling error from training error because each denoiser call returns the conditional mean under forward noising. After two calls at high initial noise, the mean absolute residual is small, yet paths with the fewest possible switches occur more often after decoding each bin than the conditional law predicts (Figure~\ref{fig:hmm}). In the solar wind experiments the denoiser is learned, and changing the loss can change both its predictions and which training step validation selects.

Other training procedures change the influence of the residual itself. Physics-Informed Learning via Diffusion (PILD) adjusts its weight at each noise level using a Laplace likelihood. The approximation needs the conditional mean and variance of residual magnitude given the noisy input and observations. PILD estimates these moments from a minibatch at each noise level and smooths the estimates across iterations \citep[Section 3.1 and Appendix A.2]{zeng2026pild}. Physics-Based Flow Matching integrates to $t=1$ before evaluating the residual. It combines the gradients of the flow matching loss and the residual loss to find a descent direction for both \citep{baldan2025physics}. The solar wind coefficient sweep instead changes the strength of the same magnetic penalty at the same point in the denoising procedure.

Algebraic completion can act on an existing forecast or change what the model learns. \citet{beucler2021constraints} compare training on all outputs, including those calculated by a conservation layer, with training on only the directly predicted outputs. In the magnetic forecasts, deriving vector variance after sampling leaves the five scored quantities untouched. Training G5 also removes vector variance from the network outputs and data loss, so the five scored forecasts can change. In a Gaussian variational autoencoder, adding the same fraction of each input to its reconstruction can lower reconstruction mean squared error while preserving the law of generated observations if the decoder output and residual noise variance are rescaled together \citep[Proposition~1 and Section~4.1]{lee2026inputskip}. In separate fits to synthetic Gaussian data, adding the skip and increasing observation variance together lowered median reconstruction error but raised the median energy score.

Completing each sample enforces a particular relation. For each draw of the directly forecast quantities, the bottom-up benchmark of \citet{biswas2026reconciliation} calculates the remaining quantities from the constraint. For example, deriving $v_v$ from the magnetic identity leaves any negative prediction of $v_s$ unchanged, although Equation~\ref{eq:support} requires $v_s\geq 0$. \citet{harder2023hard} likewise evaluate agreement with the coarse input after aggregation, the number of negative water content predictions, and errors relative to the fine-grid reference fields. Even projections that enforce the same conservation law can give different forecasts. When both models minimize the continuous ranked probability score (CRPS) during training, the orthogonal projection gives the lower test CRPS for the heat equation, and the oblique projection gives the lower test CRPS for the Stefan problem \citep[Table~2]{utkarsh2026hard}.

The observation rule determines which relation to impose. For temperatures in kelvins, averaging fourth powers and taking the fourth root under constant emissivity can give a different coarse temperature from the arithmetic mean. \citet{zhang2025pgdm} compare the two averaging rules. After training a diffusion model with geophysical inputs and a mean squared error loss, they aggregate its predictions back to the coarse scale and compare them with the original temperatures. Weather Inverse Diffusion (WIND) uses spatial averages to reconstruct atmospheric fields on a finer grid and daily means to reconstruct fields every six hours. After pretraining with diffusion forcing, it approximates the clean state given the noisy state by a Gaussian and propagates its covariance through a local linearization of the observation operator \citep{aich2026wind}.

An iterative variational autoencoder flagged possible anomalies at a NOAA DART station by thresholding the absolute difference between observed and reconstructed ocean-bottom pressure \citep[Section~4.2.3]{lee2024iterative}. For one component of its anomaly score, PISCES averages the absolute discrepancy between squared reconstructed field magnitude and squared vector norm over the minutes in an hour. PISCES tests how event detection changes when both the magnitude and Parker spiral penalties are added to the reconstruction loss \citep[Table~3]{lee2026pisces}. During training, PISCES divides each minute's residual by the larger of the squared reconstructed magnitude and $1\,\mathrm{nT}^2$, then squares the ratio. I-S takes the same difference in squared field strengths after averaging over five minutes. The predicted moments inside the gap suffice to calculate both the I-S and O-S residuals. The two methods use the same training scale $B_{\mathrm{ref}}^2$, coefficient, and estimator S. O-S also subtracts the difference between vector variance and variance of field magnitude.

\citet{kondrashov2010gaps} reconstruct hourly solar wind dynamic pressure and the north--south magnetic component from their covariation with geomagnetic indices. Separate Gaussian process regressions estimate each upstream plasma or magnetic quantity between visits of the Mars Atmosphere and Volatile Evolution (MAVEN) spacecraft to the upstream solar wind \citep{azari2024virtual}. Estimating second-order structure functions from incomplete magnetic records \citep{wrench2022statistics} and forecasting future extreme ultraviolet images of the solar corona \citep{francisco2024corona} pose different prediction problems. Here the forecast describes how the magnetic moments vary together across quantities and bins. Within each bin the moments retain variation among the minute averages, but permuting those averages leaves the moments unchanged.

\FloatBarrier
\section{Reconstruction examples}
\label{app:imputation}

\begin{figure}[p]
 \centering
 \includegraphics[width=0.95\linewidth,trim=0 6bp 0 0]{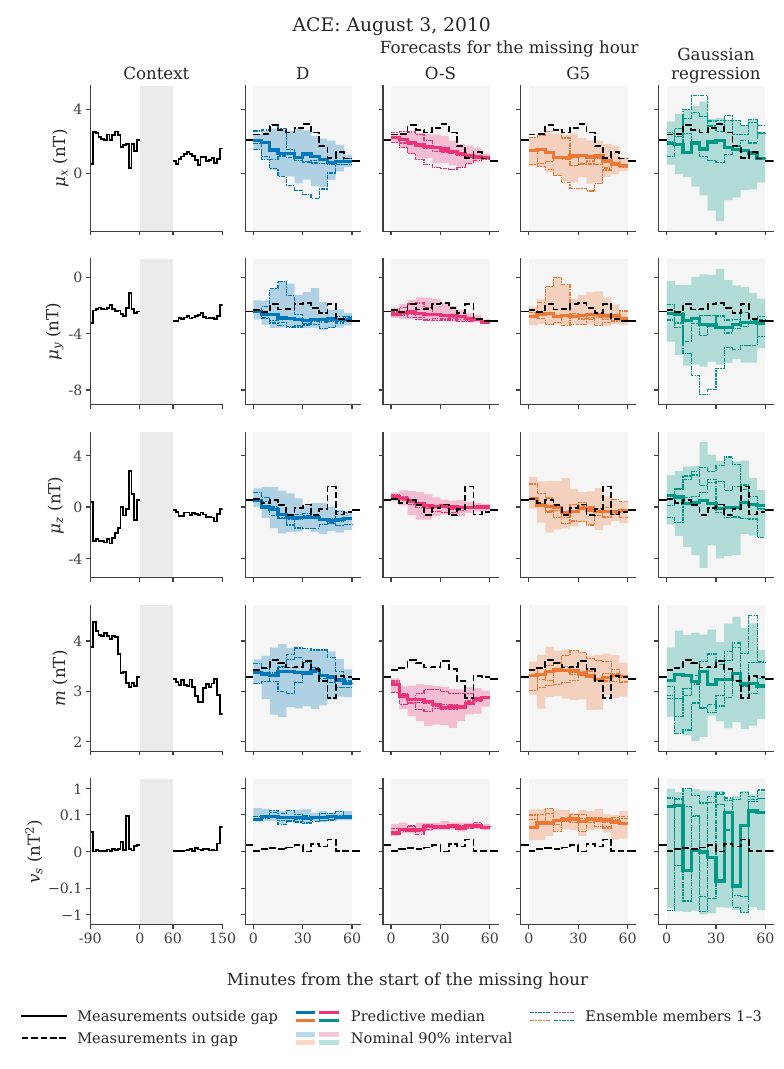}
 \caption{For the Advanced Composition Explorer (ACE) on August 3, 2010, the gap in test window 120 spans 13:30--14:30 Coordinated Universal Time on the OMNI grid of arrival times. The left column shows 90 minutes of input on each side. In the D, O-S, G5, and Gaussian regression columns, solid black steps mark the five-minute input bins beside the shaded hour. Within each row, all methods use the same vertical limits. The mean field components and mean field magnitude are in nT. The variance of field magnitude $v_s$ is in nT$^2$. Dashed black steps show the measurements in the gap, excluded from the inputs. Solid colored steps give the predictive median (rank 20) and bands give the nominal 90\% interval (ranks 2--38) of each method's 39 ensemble members at each bin and coordinate. Thin steps trace each model's first three stored forecasts for the whole gap. Each step spans five minutes. Validation energy scores for raw forecasts select the diffusion training steps and the Gaussian regression parameters. The raw samples use diffusion training seed 7 and Gaussian regression sampling seed 13. The $v_s$ scale is linear within $0.05$ nT$^2$ of zero and logarithmic outside.}
 \label{fig:imputation120}
\end{figure}

\begin{figure}[p]
 \centering
 \includegraphics[width=0.95\linewidth]{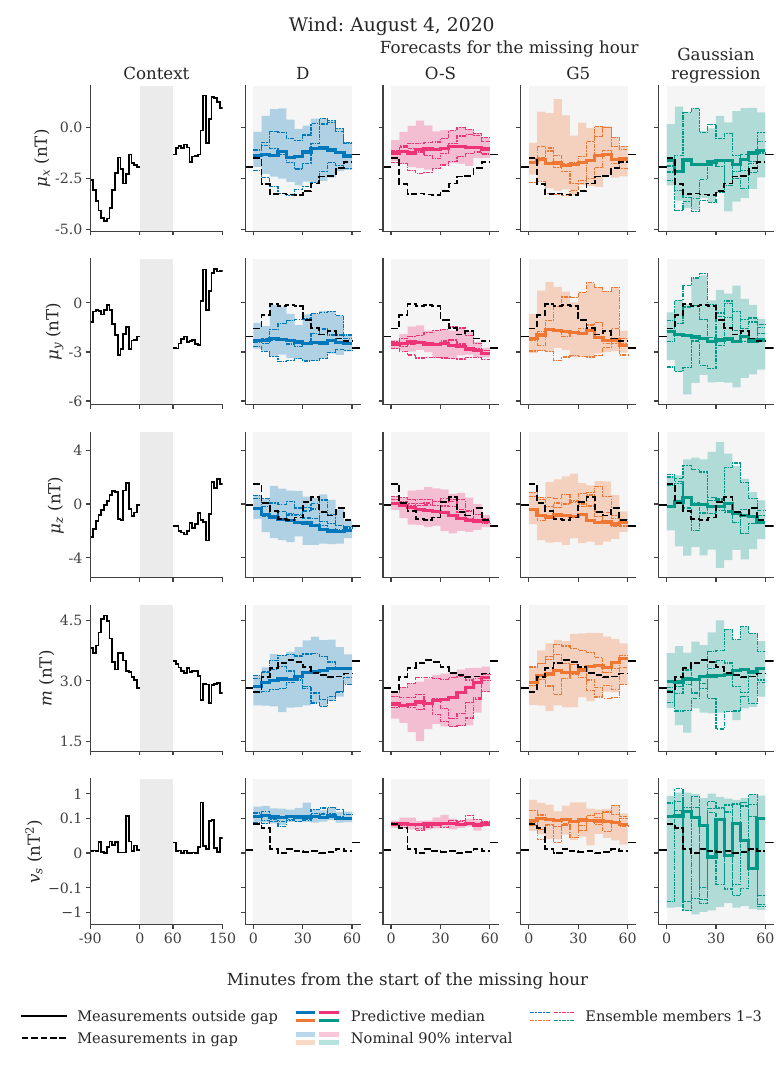}
 \caption{For Wind on August 4, 2020, the gap in test window 360 spans 18:30--19:30 Coordinated Universal Time on the OMNI grid of arrival times. The raw forecasts use diffusion training seed 7 and Gaussian regression sampling seed 13. The left column shows 90 minutes of input measurements before the shaded gap and 90 minutes after it. Each forecast column shows the missing hour in steps of five minutes. Solid black steps show the input bins just before and after the shaded hour. Dashed black steps show the measurements in the gap, excluded from the inputs. Solid colored steps mark the predictive medians, bands span ranks 2--38 of 39, and thin steps trace each model's first three stored forecasts for the whole gap. The bands give a nominal 90\% interval at each bin and coordinate. The scale for the variance of field magnitude $v_s$ is linear within $0.05$ nT$^2$ of zero and logarithmic outside.}
 \label{fig:imputation360}
\end{figure}

Figures~\ref{fig:imputation120} and \ref{fig:imputation360} show how the same forecast can succeed for one magnetic quantity and miss another. In both hours, the intervals from unpenalized diffusion cover the measured mean field magnitude in every bin and the variance of field magnitude in none. During the Wind gap, the mean $x$ component stays below $-3$ nT for five consecutive bins. All four predictive medians stay above $-2$ nT throughout those bins. In the ACE gap, the mean $z$ component changes sign five times, while the standard deviation of field magnitude within each bin stays below 0.15 nT throughout the hour. Changes in field direction across bins can coexist with little variation in field strength within them.

Taking the median separately in each bin and coordinate can combine values from different ensemble members. Thin steps trace the first three stored forecasts from each model across all twelve bins of the gap. Each forecast fills the gap and retains the measurements on both sides. At either edge of the gap, the solid and dashed black steps show measurements from adjacent bins, each five minutes wide. Their heights can differ even in the measured record. Compare the median with the measurement in the same bin of the gap to see where it misses.

The examples come from the 121st window in each half of the ordered TEST list (indices 120 and 360 in the full list, counting from zero). The choice of positions followed earlier analyses of TEST. How well the two windows represent the full TEST set is unknown. Each diffusion fit uses the training step with the lowest energy score for raw forecasts on the validation windows. Gaussian regression uses the combination of ridge coefficient and covariance shrinkage with the lowest such score.

\begin{figure}[ht]
 \centering
 \includegraphics[width=\linewidth]{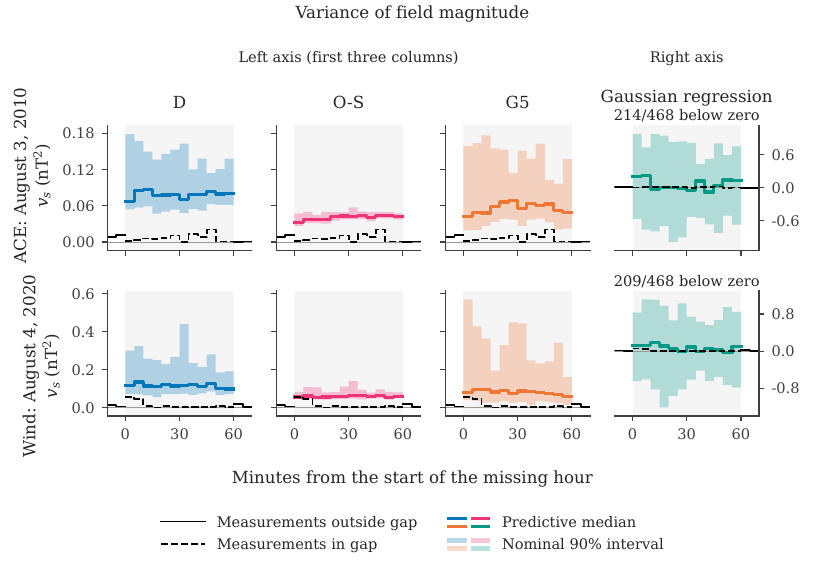}
 \caption{Variance of field magnitude $v_s$ in the same two missing hours, before clipping. Each measured value is the population variance of the magnitudes of five vectors, each averaged over one minute. D, O-S, and G5 share a linear scale within each row. The separate Gaussian regression scale accommodates its wider intervals. The limits on each axis include every measurement in the time range and all medians and interval endpoints for the methods on that axis. Gray shading marks the missing hour from 0 to 60 minutes. Solid black lines show input measurements outside the gap, and dashed lines show measurements inside it. Colored steps give the predictive median in each bin, and bands span the second and 38th ordered values among the 39 members of each ensemble, defining inclusive nominal 90\% intervals. Each step covers a full bin of five minutes. Values below the gray zero line are physically impossible.}
 \label{fig:imputationdetail}
\end{figure}

\clearpage
Figure~\ref{fig:imputationdetail} shows where the forecasts of variance of field magnitude fail. Each measured variance in the gap is calculated across five magnitudes, one for each minute's mean field vector. Here the interval width is the difference between the 38th and second ordered predictions for $v_s$ in a bin. In both examples, whenever a diffusion interval misses, the measured variance falls below its lower endpoint. For window 120, the three diffusion forecasts each cover zero of the twelve hidden $v_s$ values. For window 360, D, O-S, and G5 cover zero, one, and two bins, respectively. Across TEST, the raw intervals for $v_s$ also fall short of 90\% coverage for all three methods (Table~\ref{tab:channelcoverage}). The cause of the high diffusion medians in these two windows is uncertain.

Gaussian regression's much wider intervals contain all twelve measured $v_s$ values in each gap. Its residual covariance is the same for every context, and its predictive distribution gives positive probability to $v_s<0$. Each label above the regression plots gives the number below zero out of 468 predictions (39 members in each of twelve bins). Nearly half the Gaussian variance predictions in each window fall below zero.

\end{document}